\documentclass[a4paper, 11pt]{article}

\usepackage[mono=false]{libertine}
\usepackage[top=2.5cm, bottom=2.5cm, left=2cm, right=2cm]{geometry}

\RequirePackage{algorithm}
\RequirePackage{algorithmic}
\usepackage{amsfonts}       
\usepackage{amsmath}
\usepackage{amssymb}
\usepackage{authblk}
\usepackage{bbm}
\usepackage{multirow}

\renewcommand{\vec}[1]{\boldsymbol{#1}}
\def\mat#1{\text{#1}}

\newcommand{\fcp}{\rm fcp}

\newcommand{\dataset}{\mathcal{D}}
\newcommand{\quantile}{\hat{q}}
\newcommand{\score}{\sigma}
\newcommand{\coverage}{\kappa}

\newcommand{\gamp}{\rm gamp}
\newcommand{\amp}{\rm amp}
\newcommand{\channel}{g_{\rm out}}
\newcommand{\denoiser}{f_w}

\newcommand{\dd}{{\rm d}}

\newcommand{\bo}{\rm bo}

\newcommand{\interval}{\mathcal{S}}
\newcommand{\jaccard}{\mathcal{J}}

\newcommand{\vectheta}{\vec{\theta}}
\newcommand{\wstar}{\Vec{\theta}_{\star}{}}
\newcommand{\what}{\hat{\Vec{\theta}}}
\newcommand{\vhat}{\hat{\Vec{v}}}

\newcommand{\taylorgamp}{Taylor-AMP }

\newcommand{\fgamp}{\vec{f}_{\gamp}} 

\usepackage[utf8]{inputenc} 
\usepackage[T1]{fontenc}    
\usepackage{hyperref}
\hypersetup{ colorlinks, breaklinks=true, urlcolor=orange, linkcolor=blue, citecolor=blue }

\usepackage[capitalise]{cleveref}
\crefname{equation}{Equation}{Equations}
\crefname{figure}{Figure}{Figures}
\crefname{table}{Table}{Tables}
    
\usepackage{url}            
\usepackage{booktabs}       
\usepackage{microtype}      
\usepackage{nicefrac}       
\usepackage{xcolor}         
\usepackage{xfrac}
\usepackage{multirow}

\usepackage{natbib}
\newtheorem{property}{Property}

\title{Building Conformal Prediction Intervals \\ with Approximate Message Passing}

\author[1]{Lucas Clart\'e}
\author[1]{Lenka Zdeborov\'a}

\affil[1]{
\'Ecole Polytechnique F\'ed\'erale de Lausanne (EPFL)\\
Statistical Physics of Computation laboratory\\
CH-1015 Lausanne, Switzerland
}
\begin{document}

\maketitle

\begin{abstract}   
    Conformal prediction has emerged as a powerful tool for building prediction intervals that are valid in a distribution-free way. However, its evaluation may be computationally costly, especially in the high-dimensional setting where the dimensionality and sample sizes are both large and of comparable magnitudes. To address this challenge in the context of generalized linear regression, we propose a novel algorithm based on Approximate Message Passing (AMP) to accelerate the computation of prediction intervals using full conformal prediction, by approximating the computation of conformity scores. Our work bridges a gap between modern uncertainty quantification techniques and tools for high-dimensional problems involving the AMP algorithm. We evaluate our method on both synthetic and real data, and show that it produces prediction intervals that are close to the baseline methods, while being orders of magnitude faster. Additionally, in the high-dimensional limit and under assumptions on the data distribution, {the conformity scores computed by AMP converge to the one computed exactly}, which allows theoretical study and benchmarking of conformal methods in high dimensions.
\end{abstract}
\section{Introduction}
\vspace{-3mm}
Quantifying uncertainty is a central task in statistics, especially in sensitive applications. For regression tasks, the goal is to produce prediction sets instead of point estimates: consider here a dataset~$\dataset = \left( (\Vec{x}_i, y_i) \right)_{i = 1}^n$ with independent samples of the same distribution, with $\left( \Vec{x}, y \right) \in \mathbb{R}^d \times \mathbb{R}$. Given a new input $\Vec{x}$, we aim to produce a set of prediction $\interval(\Vec{x})$ that contains the observed label $y$ with probability $1 - \coverage$ for $\coverage \in (0, 1)$. Conformal methods constitute a general framework used to produce such prediction sets with guarantees on their coverage. Among these methods, we can cite full and split conformal prediction (FCP and SCP) \cite{vovk1005algorithmic, shafer2007tutorial} and Jackknife+ \cite{Barber2019Predictive}. In full conformal prediction, the prediction set of $\Vec{x}$ is the set of labels $y$ whose \textit{typicalness} is sufficiently high. The computation of this typicalness is based on leave-one-out residuals that are computed on an augmented dataset that includes the test data. Full conformal prediction has been shown to provide the correct coverage under the exchangeability of the data samples and symmetry of the scoring function under the permutation of the data. However, the computation cost of FCP is proportional to the number of training samples and the number of possible labels, making it computationally very heavy in practice. Split conformal prediction (SCP) \cite{shafer2007tutorial, Lei2018Distribution} is an efficient alternative to FCP, in which data is split between training and validation sets, the latter being used to calibrate the model after training. SCP is much more efficient than FCP, at the expense of statistical efficiency. Indeed, because the model is fitted on a lower amount of data than in FCP, the intervals of SCP are wider and thus less informative than FCP, as illustrated in~\cite{Lei2018Distribution}. Other works are concerned with accelerating full conformal prediction. For instance, \cite{lei2017fast, ndiaye2019computing} approximate the computations of FCP by linearizing the solution of an empirical risk minimization problem. While \cite{lei2017fast} is limited to the Lasso, \cite{ndiaye2019computing} is applicable to general convex empirical risks. Additionally, the work of \cite{cherubin2021exact} leverages incremental learning in the context of classification, kernel density estimation and k-NN regression.

\vspace{-1em}
\paragraph{Uncertainty quantification in high dimensions --} In this work, we will focus our attention on the \textit{high-dimensional} regime, where the number of samples $n$ and the dimension $d$ are both large with a fixed ratio $\alpha = \sfrac{n}{d}$. In this regime, many common uncertainty quantification methods are not applicable or quantify the true uncertainty wrongly. Full conformal prediction is computationally demanding as it needs to fit $n$ estimators for each possible label. Alternatives, such as split conformal prediction or the Jackknife+~\cite{Barber2019Predictive} are more tractable, at the expense of statistical efficiency.
On the other hand, the bootstrap~\cite{Davison1997Bootstrap} has been shown to fail in high-dimensional linear regression~\cite{clarte2024analysis, Karoui2018Can} and with deep neural networks~\cite{Nixon2020WhyA}. Other methods based on ensembling, like the jackknife~\cite{Quenouille1956Notes} or Adaboost~\cite{Zhu2006Multiclass}, have been analyzed in high-dimension~\cite{takahashi2024replica, clarte2024analysis, Loureiro2023Fluctuations, Liang2022Precise} and have been shown to be problematic in that setting as well. Authors of \cite{Bai2021Understanding} have shown that unpenalized quantile regression achieves under-coverage in high dimensions. 
\paragraph{High-dimensional inference with AMP --} Approximate message passing (AMP) algorithms are a class of iterative equations used to solve inference problems in high-dimension under certain distributional assumptions~\cite{Donoho2009Message,Zdeborova2016Statistical}. These equations are usually derived by relaxing belief propagation equations in a graphical model~\cite{Pearl2014Probabilistic}. A central property of AMP algorithms is their state-evolution equations that track their behaviour in high dimensions. Thanks to these state-evolution equations, AMP has been used as an analytical tool to tackle a wide range of problems in high-dimensional statistics \cite{Sur2019Modern,Donoho2009Message,Bayati2010Dynamics}. In the context of uncertainty quantification, AMP has been used to study the calibration of frequentist and Bayesian classifiers~\cite{Bai2021Dont, clarte2023theoretical, clarte2022overparametrized} and for change point detection \cite{arpino24Inferring}. Additionally to these analyses, AMP algorithms have also been used in practical scenarios, such as compressed sensing~\cite{Donoho2009Message}, genomics~\cite{Depope2024Inference}, to accelerate cross-validation \cite{Obuchi2016Cross} or for change point detection~\cite{arpino2024inferring}. Finally, in Bayesian learning, AMP can be used to compute marginals of the posterior distributions faster than with Monte-Carlo methods \cite{clarte2023theoretical}, or it can be used to establish fast sampling rigorously \cite{el2022sampling}. However, to our knowledge, no work has applied AMP to accelerate the computation of full conformal prediction.
\paragraph{Contributions --} Our contributions are three-fold:
\vspace{-3mm}
\begin{itemize}
    \item First, we apply the AMP algorithm on generalized linear regression to compute the prediction intervals of full conformal prediction. AMP accelerates FCP by approximating the $n$ leave-one-out estimators simultaneously. We show that it still provides coverage guarantees under the standard assumption that the data is exchangeable.
    \item Second, we introduce the \taylorgamp algorithm, which further accelerates the computations by removing the need to fit an estimator for each possible label. We claim that \taylorgamp is a good approximation of AMP if the empirical risk minimizer only weakly depends on each sample.
    \item Finally, we show that in a teacher-student model with Gaussian data and in the high-dimensional limit, AMP recovers the prediction intervals obtained by computing the leave-one-out scores exactly. As a consequence, our algorithm allows the study of conformal prediction in high dimensions and provides a non-trivial benchmark for other methods in this regime. We also demonstrate the performance of \taylorgamp on real data.
\end{itemize}

To our knowledge, our work is the first to apply ideas from the area of approximate message-passing algorithms to full conformal prediction and opens the door to a new research direction in which methods from high-dimensional statistics can be used practically for uncertainty quantification. The AMP-based method has the coverage guarantees celebrated in conformal prediction, with possible wide prediction intervals if the scores are estimated inacurately. The method can be used with practical advantages in scenarios where the AMP is usable for estimation, for instance, genomics \cite{Depope2024Inference} or MRI reconstruction~\cite{Millard2020MRI}. Another practical interest of our work stems from the utility of having non-trivial high-dimensional settings where FCP can be evaluated rapidly, as this may be useful for theoretical research and benchmarking of other more general speed-up methods.  
\paragraph{Notation --} For a set of real values $\Vec{z} = z_1, \cdots, z_n$ we will write $\quantile_{\coverage} \left( \Vec{z} \right)$ the $\coverage$ quantile of $\Vec{z}$ (i.e the $\kappa \times n$ largest value). The normal distribution of mean $\mu$ and variance $\sigma^2$ will be noted $\mathcal{N}(\mu, \sigma^2)$ while we will denote by $\mathcal{L}(\mu, b)$ the Laplace distribution with density $p(x) = \frac{1}{2b} e^{-\frac{|x - \mu|}{b}}$. The element-wise product between two vectors or matrices $A, B$ will be written $A \otimes B$. ${\rm Jac}$ denotes the Jacobian of a vector-valued function.

\vspace{-3mm}
\section{Setting}
\vspace{-3mm}
\label{sec:setting}


We consider here the framework of generalized linear models for regression. Assume a training set $\mathcal{D} = \left( \Vec{x}_i, y_i \right)_{i = 1}^{n}$ where the $n$ samples $\Vec{x}_i, y_i \in \mathbb{R}^d \times \mathbb{R}$ are i.i.d. Given a test sample~$\Vec{x}$, we want to build a prediction set $\interval(\Vec{x})$ that contains the true label $y$ with probability $1 - \coverage$
\begin{equation}
    \mathbb{P}_{\dataset, \Vec{x}} \left( y \in \interval(\Vec{x}) \right) \geqslant 1 - \coverage \, .
    \label{eq:def_coverage}
\end{equation}
In \eqref{eq:def_coverage}, the randomness is on the training data and the test sample.
We are interested in methods that provide the correct coverage with prediction sets of minimal size. 
In this work, we will focus on generalized linear models trained using empirical risk minimization 
\begin{equation}
    \what = \arg\min_{\Vec{\theta}} \mathcal{R} \left( \Vec{\theta} \right) = \arg\min_{\Vec{\theta}} \sum_{i = 1}^n \ell \left( y_i, \Vec{\theta}^{\top} \Vec{x}_i \right) + \sum_{\mu = 1}^d r(\theta_{\mu})
    \label{eq:def_erm}
\end{equation}
where $\ell$ is a convex loss and $r$ is a convex regularizer. For concreteness, we will consider the cases of Ridge ($r(\theta) = \frac{\lambda}{2} \theta^2$) and Lasso ($r(\theta) = \lambda |\theta |$) regression, but our results are easily extendable to other problems such as quantile regression. Because the methods that we investigate rely on the computation of leave-one-out residuals, we introduce the leave-one out estimators $\what_{-i}$ that are learned on the whole dataset except sample $i$

\subsection{Full conformal prediction} 
\vspace{-2mm}
The basic procedure of full conformal prediction is to iterate over any possible label $y$, for which we define the augmented dataset $\dataset^+ \left( y \right) = \dataset \cup (\Vec{x}, y)$. We then compute the $n+1$ leave-one-out estimators $\what_{-i}$ trained on $\dataset^+ \left( y \right)$ from which we compute the conformity scores $\score_i \left( y \right)$. These scores will be used to compute test statistics that will determine the inclusion $y$ in the prediction set $\interval \left( \vec{x} \right) $. We first define 
\begin{align}
    \label{eq:argmin_loo}
    \what_{-i} \left( y \right) &= \arg\min_{\vec{\theta}} \sum_{j \neq i } \ell \left( y_j, \Vec{\theta}^{\top} \Vec{x}_j \right) + \ell \left( y, \Vec{\theta}^{\top} \Vec{x} \right)\\ &+ \sum_{\mu} r \left( \theta_{\mu} \right)\nonumber
\end{align}
that minimizes the empirical risk on $\dataset^+ \left( y \right)$. We then define the conformity scores as the leave-one-out residuals:
\begin{equation}
    \score_i(y) = | \what_{-i} \left( y \right)^{\top} \Vec{x}_i - y_i | 
    \label{eq:scores_fcp}
\end{equation}
From these scores, the prediction set $\interval_{\fcp} \left( \Vec{x} \right)$ is defined by 
\begin{equation}
\label{eq:def_fcp_interval}
    y \in \interval_{\fcp} \left( \Vec{x} \right) \Leftrightarrow \score_{n+1} \left( y \right) \leqslant \quantile_{ \lceil \left(1 - \coverage \right) \left( n + 1 \right) \rceil / n} (\Vec{\score}(y))
\end{equation}
in other words, a label $y$ is included in the prediction set if the conformity score of the test sample, when using the $y_{n+1} = y$, is lower than the $\sfrac{\lceil \left( 1 - \coverage \right) (n + 1) \rceil}{n} $ quantile of the scores $\score_1(y), \cdots, \score_{n+1}(y)$. \cite{vovk1005algorithmic, angelopoulos2022gentle}.

In what follows, we will refer as \textit{exact LOO} the computation of the conformity scores~\eqref{eq:scores_fcp} by solving the minimization problems~\eqref{eq:argmin_loo} exactly. The prediction set $\interval_{\fcp}$ achieves the desired coverage on average under the assumption that the data is exchangeable and the regression function used to produce the conformity scores is symmetric \cite{vovk1005algorithmic}. However, as noted before, fitting a model for all possible labels and computing the residuals by solving the minimization problem~\eqref{eq:argmin_loo} is computationally heavy in practice. Methods have been developed to accelerate the computation of full conformal prediction, and in this paper, we introduce two algorithms that leverage tools from high-dimensional statistics, namely the AMP and Taylor-AMP algorithms. Contrary to exact LOO, our methods approximate the computation of the leave-one-out estimators~\eqref{eq:argmin_loo} used to build prediction intervals.

\subsection{Split conformal prediction} 
\vspace{-2mm}
Split conformal prediction (SCP, also known as inductive conformal prediction)~\cite{Papadopoulos2002Inductive, vovk1005algorithmic} is an alternative to FCP that is computationally much cheaper. In the simplest form of SCP, $\dataset$ is split between the training and calibration sets $\dataset_{\rm train}, \dataset_{\rm cal}$. An estimator $\what$ will be fit using $\dataset_{\rm train}$, and the conformity scores $\left( \score_i \right)_{i = 1}^{|\dataset_{\rm cal}|}$ are computed on the calibration set. We then extract the $\lceil ( 1 - \coverage ) \times (n + 1) \rceil$ quantile of the scores. 
\begin{align}
    \score_i = | y_i - \what^{\top} \Vec{x}_i |, &\qquad Q = \quantile_{ \sfrac{\lceil ( 1 - \coverage ) \times (n + 1) \rceil}{n
    } } \left( \score_i \right) \\
    \interval_{\rm SCP} \left( \Vec{x} \right) &= \left[ \what^{\top} \Vec{x} - Q, \what^{\top} \Vec{x} + Q\right]
    \label{eq:simple_scp}
\end{align}

One drawback of \eqref{eq:simple_scp} is that its prediction intervals are of the same size for all test samples. In this context, \cite{Romano2019Conformalized} introduced conformal quantile regression, which combines split conformal prediction and quantile regression to accommodate potential heteroskedasticity and produce intervals with data-dependent length. 

\subsection{Bayes-optimal estimator}
\vspace{-2mm}
\label{sec:def_bayes_optimal}
Consider the Bayesian setting where the parameter to infer $\wstar$ is sampled from a prior $p_{\wstar}$ and the labels are generated by the likelihood distribution $p(y | \wstar^{\top} \Vec{x})$. One can then compute the Bayes posterior 
\begin{equation}
    \Vec{\theta} \sim p( \vec{\theta} | \dataset ) \propto \prod_{i = 1}^n p \left( y_i | \wstar^{\top} \Vec{x}_i \right)  p_{\wstar} \left( \wstar \right)
\end{equation}
which yields the \textit{Bayes-optimal} estimator, with the lowest generalisation error. This posterior distribution yields the predictive posterior distribution 
\begin{equation}
    p( y | \dataset, \Vec{x}) = \int \dd \Vec{\theta} p(y | \Vec{\theta}^{\top} \Vec{x}) p(\Vec{\theta} | \dataset )
\end{equation}
One can then build a prediction interval $\interval_{\bo} ( \Vec{x} )$ for the Bayes-optimal estimator using the \textit{highest density interval}, which for a coverage $1 - \coverage$ is the smallest set with measure $1 - \coverage$.

\vspace{-1em}
\paragraph{Bayes posterior and maximum a posteriori} In some settings, the empirical risk~\eqref{eq:def_erm} corresponds to the logarithm of the Bayes-posterior. For instance, Ridge regression with $\lambda = 1$ corresponds to the log-posterior for the Gaussian prior $p_{\wstar} = \mathcal{N}(0, 1)$ while Lasso with $\lambda = 1$ matches the log posterior for the Laplace prior $p_{\wstar} = \mathcal{L}(0, 1)$.

\section{Approximate message passing for uncertainty quantification}
\label{sec:amp_for_uncertainty}

\subsection{Computing residuals using AMP}

We first introduce the AMP algorithm, stated in Algorithm~\ref{alg:gamp}. Given the regression problem~\eqref{eq:def_erm}, AMP approximates $\what_{\gamp}$ of the empirical risk minimizer $\what$. As we will show later, using AMP to solve~\cref{eq:def_erm} will allow us to simultaneously compute all the leave-one-out estimators instead of fitting the model $n$ times, thus dramatically accelerating the computations. 
While AMP has been discussed extensively in the literature, for example, in \cite{Donoho2009Message,Zdeborova2016Statistical,Mezard2009Information}, we point the reader to Appendix~\ref{appendix:amp} for its derivation.

Algorithm~\ref{alg:gamp} requires to define a \textit{channel} and \textit{denoising} functions, respectively noted as $\channel$ and $\denoiser$ and defined as follows depending on the choice of loss and regularization:
\begin{align}
    \channel(y, \omega, V) &= \arg\min_z \ell \left( z, y \right) + \frac{1}{2V} \left( z - \omega \right)^2\\ \denoiser(b, A) &= \arg\min_z r(z) + \frac{1}{2A} (z - A b)^2
    \label{eq:def_channel_denoiser}
\end{align}
Above, $\channel$ and $\denoiser$ take scalar arguments but are applied on vectors in Algorithm~\ref{alg:gamp} by applying the functions component-wise.

\paragraph{Channel and denoiser for Ridge and Lasso -- } In the general setting, computing $\channel$ and $\denoiser$ requires minimizing a scalar function. In this work, we will focus on Ridge regression and the Lasso, where these functions have a closed-form expression
\begin{align}
&\begin{cases}
    \channel^{\rm Ridge}(y, \omega, V) \!\!\! &= \frac{y - \omega}{1 + V} \\
    \denoiser^{\rm Ridge}(b, A) \!\!\!&= \frac{b}{\lambda + A}
\end{cases}, \\
&\begin{cases}
    \channel^{\rm Lasso}(y, \omega, V) \!\!\!&= \frac{y - \omega}{1 + V} \\
    \denoiser^{\rm Lasso}(b, A) \!\!\!&= \frac{b - \lambda}{A} \text{ if } b > \lambda, \frac{b + \lambda}{A} \text{ if } b < - \lambda  \, \text{else} \, \,  0  \nonumber
\end{cases}
\end{align}

\paragraph{Leave-one-out estimation --} Using AMP, one can approximate the leave-one-out-estimators~\eqref{eq:argmin_loo} and the associated residuals~\eqref{eq:scores_fcp} with a single fit of the algorithm: for any sample $i$, an approximation of the $\what_{-i}$ is given by the following expression 
\begin{equation}
    \what_{-i, \gamp} \left( y \right) = \what_{\gamp} \left( y \right) - g_{i, \gamp} \left( y \right) \times \Vec{x}_{i}^{\top} \otimes \vhat_{\gamp} \left( y \right)
    \label{eq:leave_one_out_from_amp}
\end{equation}
where all the vectors $\what_{\gamp}, \vhat_{\gamp}, \Vec{g}_{\gamp}$ are computed in Algorithm~\ref{alg:gamp}, and the dependency on the last label $y$ is made explicit. We refer the reader to Appendix~\ref{appendix:amp} for a justification of the above expression. The derivation is based on a close cousin of AMP, relaxed Belief Propagation (rBP), which is equivalent in the high-dimensional limit under Gaussianity assumptions on the data distribution, which we discuss in~\cref{sec:convergence_high_dim}. At finite dimensions $d$ the leave-one-out estimators $\what_{-i,\gamp}$ from \eqref{eq:leave_one_out_from_amp} are only approximations of the solutions of~\eqref{eq:argmin_loo} and may not be very good approximations. However, they still provide valid coverage guarantees, as essential in the conformal prediction. 

\begin{algorithm}[tb]
    \caption{AMP}
    \label{alg:gamp}
    \begin{algorithmic}
        \STATE {\bfseries Input:} Dataset $\dataset = \left( \Vec{x}_i, y_i \right)_{i = 1}^n$ 

        \STATE \hrule
        
        \STATE Define $\mat{X}^2 = \mat{X}\otimes \mat{X} \in\mathbb{R}^{n\times d}$ and initialize $\what^{t=0} = \mathcal{N}(\vec{0}, \mat{I}_{d})$, $\hat{\vec{v}}^{t=0} = \vec{1}_{d}$, $\vec{g}^{t=0} = \vec{0}_{n}$.
        \FOR{$t\leq t_{\text{max}}$ or until convergence}
            \STATE $\vec{V}^{t} = \mat{X}^{2} \hat{\vec{v}}^{t}$ ; $\vec{\omega}^{t} = \mat{X} \what^{t} - \vec{V}^{t}\otimes \vec{g}^{t-1}$ ; \qquad\textit{/* Update channel mean and variance}
            \STATE $\vec{g}^{t} = \channel(\vec{y}, \vec{\omega}^{t}, \vec{V}^{t})$ ; $\partial\vec{g}^{t} = \partial_{\omega} \channel(\vec{y}, \vec{\omega}^{t}, \vec{V}^{t})$ ; \qquad\textit{/* Update channel}
            \STATE $\vec{A}^{t} = -{\mat{X}^{2}}^{\top} \partial \vec{g}^{t}$ ; $\vec{b}^{t} = \mat{X}^{\top} \vec{g}^{t} + \vec{A}^{t}\otimes \what^{t}$ ; \qquad\textit{/* Update prior mean and variance } 
            \STATE \textit{/* Update marginals */}
            \STATE $\what^{t+1} = \denoiser (\vec{b}^{t}, \vec{A}^{t}) $ ;\qquad $\hat{\vec{v}}^{t+1} = \partial_{b} \denoiser (\vec{b}^{t}, \vec{A}^{t})$ 
        \ENDFOR

        \STATE\textit{/* Compute the leave-one-out estimators with~\cref{eq:leave_one_out_from_amp}}
        \FOR{$1 \leqslant i \leqslant n$} 
            \STATE $\what_{-i, \gamp} = \what_{\gamp} - {g}_{{\gamp},i} \Vec{x}_i \otimes \hat{\Vec{v}}_{\gamp}$
        \ENDFOR
        
        \STATE {\bfseries Return:} $\what_{\gamp}, ( \what_{-i, \gamp} )_{i = 1}^n$
    \end{algorithmic}
\end{algorithm}

\paragraph{Coverage guarantees for AMP --}

A central property of conformal prediction is that under very weak assumptions, one get prediction sets that have the correct coverage. Indeed, a standard property of FCP is that if the data is exchangeable and the score function $f$, which maps samples to confirmity scores, is symmetric, then the prediction intervals given by $f$ satisfy \cref{eq:def_coverage}, as shown in \cite{vovk1005algorithmic}. Recall that \textit{symmetric} means here that for any permutation $s: [1, n] \to [1, n]$, then $\hat{f}( \left( \vec{x}_{s(i)}, y_{s(i)} \right) )_{i = 1}^n = \left( \score_{s(i)} \right)_{i = 1}^n$. We show in Appendix~\cref{appendix:amp_coverage} that AMP is symmetric, which leads to the following property: 

\begin{property}
    Consider training data $\mathcal{D} = \left( \vec{x}_i, y_i \right)_{i = 1}^n$ and a test sample~$\Vec{x}$ where the data is exchangeable. Consider the conformity scores $\left( \score_{i, \gamp} \right)_i = |y_i - \what_{-i, \gamp}^{\top} \Vec{x}_i |$ where the leave-one-out estimators are computed using AMP:
$$
\what_{-i, \gamp} = \what_{\gamp} - g_{i, \gamp} \Vec{x}_{i}^{\top} \otimes \vhat_{\gamp} 
$$
and the confidence set with target coverage $1 - \kappa$, defined as 
$$
\interval_{\fcp} ( \Vec{x} ) = \left\{ y | \score_{n+1} \leqslant \quantile_{\lceil (1 - \kappa) (n + 1) \rceil / n} \left( \score_i \right) \right\}
$$
then, $\interval_{\fcp}$ achieves coverage at $1 - \kappa$ on average
\begin{equation}
\mathbb{P}_{\dataset, \Vec{x}} \left( y \in \interval_{\fcp} \left(\Vec{x}\right) \right) \geqslant 1 - \kappa 
\end{equation}
\label{prop:coverage}
\end{property}
\vspace{-4mm}
Note that~\cref{prop:coverage} is valid at finite dimension and independently of the data distribution : AMP needs not to approximate precisely the leave-one-out residuals to achieve the correct coverage.

\subsection{\taylorgamp} In the previous paragraphs, we saw that AMP can be used to accelerate the computation of the conformity scores $\sigma_i \left( y \right)$ by computing the $n$ leave-one-out estimators simultaneously for a fixed label $y$ of the test data. In this section, we present a variant of AMP called \taylorgamp and described in Algorithm~\ref{alg:gamp_order_one}, whose goal is to further accelerate AMP by approximating the iteration over the set of possible labels: \taylorgamp will compute the leave-one out estimators $\what_{-i,\gamp}\left( y \right)$ without fitting the model for each label $y$. The general idea is to approximate the quantities $\what_{-i}^{\top} \Vec{x}_i$ by an affine function around a reference value $\hat{y}$. To do so, we will compute the derivative of the estimators $\what_{ -i } (y)$ with respect to $y$, around $\hat{y}$. Then, for any possible label $y$, the corresponding scores will be approximated with  
\begin{align*}
    &\score_i \left( y \right) = | y_i - \what_{-i, \gamp} \left( y \right)^{\top} \Vec{x}_i | \\ &=  | y_i - \left( \what_{-i, \gamp} \left( \hat{y} \right) + (y - \hat{y}) \frac{\partial \what_{-i, \gamp}}{\partial y} \left( \hat{y} \right) \right)^{\top} \Vec{x}_i |
    \label{eq:scores_taylorgamp}
\end{align*}
The central part is the estimation of $\frac{\partial \what_{-i, \gamp}}{\partial y}$ using AMP. Indeed, $\what_{\gamp}$ solves a fixed point equation of the form 
$$\fgamp \left( \what_{\gamp} \left( y_{n+1} \right), y_{n+1} \right) = \what_{\gamp} \left( y_{n+1} \right)$$
where we only make explicit its dependency $y_{n+1}$ as the rest of the training data is fixed. Using the implicit function theorem, one can compute the derivative $\frac{\partial \what_{\gamp}}{\partial y_{n+1}}$ from the implicit equation
\begin{equation}
    \frac{\partial \what_{\gamp}}{\partial y}\left( \hat{y} \right) = \left( \mathbf{I} - {\rm Jac} \left( \fgamp \right) \right)^{-1} \frac{\partial \fgamp}{\partial y} \left( \hat{y} \right)
\end{equation}
which can be solved iteratively: 
\begin{equation}
    \Delta\what^{t+1} =  {\rm Jac} \left( \fgamp \right) \left( \Delta \what^{t} \right) + \frac{\partial \fgamp}{\partial y} \left( \hat{y} \right) \, .
    \label{eq:iterative_taylor_amp}
\end{equation}
In Algorithm~\ref{alg:gamp_order_one}, we iterate~\cref{eq:iterative_taylor_amp} until convergence, at which point $\left( \Delta\what, \Delta\vhat, \Delta\vec{g}\right) = \left( \frac{\partial \what}{\partial y}, \frac{\partial \vhat}{\partial y}, \frac{\partial \vec{g}}{\partial y} \right)$. We provide more details, in particular the explicit form of the function $\fgamp$ in Appendix~\ref{appendix:taylorgamp}.

To summarize, Algorithm~\ref{alg:gamp_order_one} computes the derivatives $\Delta\what_{\gamp}, \Delta\vhat_{\gamp}, \Delta \Vec{g}_{\gamp}$ of $\what_{\gamp}, \vhat_{\gamp}, \Vec{g}_{\gamp}$ around some value $\hat{y} = \what^{\top} \Vec{x}_n$ where $\what$ minimizes~\eqref{eq:def_erm} on $\dataset$. We can then approximate the leave-one-out estimators $\what_{-i,\gamp} \left( y \right)$ by differentiating the expression of the leave-one-out estimators~\eqref{eq:leave_one_out_from_amp}, which yields
\vspace{-0.5em}
\begin{align*}
    \frac{\partial \what_{-i,\gamp}}{\partial y} \left( y \right) &= \Delta \what - 
    g_{i, \gamp} \left( \hat{y} \right) \times \Vec{x}_i \otimes \Delta \vhat_{\gamp} \\ &- \Delta g_{i, \gamp} \Vec{x}_i \otimes \vhat_{\gamp} \left( \hat{y} \right)
    \label{eq:order_one_leave_one_out}
\end{align*}
which allows us to compute the conformity scores of FCP in~\cref{eq:scores_fcp}.

\vspace{-0.5em}
\paragraph{Justification of \taylorgamp --}\taylorgamp is based on the idea that the value of the last sample only weakly affects the value of the estimator $\what_{\gamp}$. More precisely, in high-dimensions as $n, d \to \infty$, $\frac{\what_{\gamp}}{\partial y} \to 0$. This implies for instance that the data contains no outliers, whose value would induce a significant change in $\what_{\gamp}$. We refer the reader to~\cref{appendix:taylorgamp_justification} for more details: we numerically observe for synthetic Gaussian data that \taylorgamp accurately approximates the leave-one-out predictions $\what_{-i}^{\top} \Vec{x}_i$ in high dimensions.

\begin{algorithm}[tb!]
    \caption{\taylorgamp}
    \label{alg:gamp_order_one}
    \begin{algorithmic}
        \STATE {\bfseries Input:} Data $\mat{X}\in\mathbb{R}^{n\times d}$, $\vec{y}\in \mathbb{R
        }^{n}$

        \STATE \hrule

        \STATE Compute $\left( \what, \hat{\Vec{v}}, \Vec{\omega}, \Vec{V}, \Vec{A}, \Vec{b}, \Vec{g}, \vec{\partial g} \right)$ using Algorithm~\ref{alg:gamp}

        \STATE Initialize $\Delta \what^0 = \Vec{0}, \Delta \hat{\Vec{v}}^0 = \Vec{0}, \Delta \Vec{V}^0 = 0, \Delta \Vec{\omega}^0 = \Vec{0}$
        
        \FOR{$t\leq t_{\text{max}}$ or until convergence}
            \STATE $\Delta \Vec{V}^t = \mat{X}^2 \Delta \hat{\Vec{v}}^{t-1}$
            \STATE $\Delta \Vec{\omega}^{t} = X \Delta \what^{t-1} - \Delta V \otimes \Vec{g}^{t-1} - V \otimes \Delta \Vec{g}^{t-1}$
            \STATE $\Delta \Vec{g}^{t} = \partial_{\Vec{\omega}} \channel \Delta \Vec{\omega}^{t} + \partial_{\Vec{V}} \channel \Delta \Vec{V}^{t} + \left( \partial_y {\channel}_{|n} \right) \vec{e}_n $
            \STATE $\Delta \partial \Vec{g}^{t} = \partial^2_{\Vec{\omega}^2} \channel \Delta \vec{\omega}^{t} + \partial_{\Vec{V}} \partial_{\Vec{\omega}} \channel \Delta \Vec{V}^{t} + \left( \partial_y \partial_{\omega} {\channel}_{|n} \right) \vec{e}_n$
            \STATE $\Delta \Vec{A}^{t} = - X^{2\top} \Delta \partial \Vec{g}^{t}$ 
            \STATE $\Delta \Vec{b}^{t} = X^{\top} \Delta \Vec{g}^{t}$ 
            \STATE $\Delta \what^{t} = \partial_{b} f_w \Delta \Vec{b}^{t} + \partial_{A} f_w \Delta \Vec{A}^{t}$ 
            \STATE $\Delta \hat{\Vec{v}}^{t} = \partial_{b} \left( \partial_{b} f_w \right) \Delta \Vec{b}^{t} + \partial_{A} \left( \partial_{b} f_w \right) \Delta \Vec{A}^{t}$ 
        \ENDFOR
        
        \STATE {\bfseries Return:} Derivatives $\left( \Delta\what_{\gamp}, \Delta\vhat_{\gamp}, \Delta \Vec{g}, \right)$
    \end{algorithmic}
\end{algorithm}

\subsection{Exactness in high dimensions for Gaussian data}
\label{sec:convergence_high_dim}

In this section, we provide guarantees on the size of the prediction intervals using conformity scores produced by AMP in high dimensions. Suppose that the samples $(\vec{x}, y)$ follow the distribution 
\begin{equation}
    y_i \sim p(\cdot | \wstar^{\top} \Vec{x}_i), \qquad \Vec{x}_i \sim \mathcal{N}(0, \sfrac{I_d}{d})
    \label{eq:gaussian_data}
\end{equation}
for $\wstar$ \textit{teacher} vector that is to be recovered from the training data and with a likelihood function $p( \cdot | z)$ that is not known to the statistician e.g. $y = \wstar^{\top} \Vec{x} + \varepsilon$, with $\varepsilon \sim \mathcal{N}(0, 1)$. Assume also that $\wstar$ is random and its components are independently sampled from the same distribution $p_{\wstar}$. In what follow we will assume that $p_{\wstar}$ is either the standard normal $p_{\wstar} = \mathcal{N}(0, 1)$ or the Laplace distribution $p_{\wstar}(z) = \frac{1}{2} e^{-|z|}$. Then, under these assumptions on $\wstar$ and the data, in the high-dimensional limit where $n, d \to \infty$ with $\sfrac{n}{d}$ fixed, the estimator $\what_{\gamp}$ converges to the true empirical risk minimizer, provided the samples $\Vec{x}_i, y_i$ come from the distribution~\eqref{eq:gaussian_data} as shown in \cite{Zdeborova2016Statistical, Mezard2009Information, Donoho2009Message}. Thus, for any test sample $\Vec{x}$ and any $\varepsilon > 0$
\begin{equation}
    \mathbb{P}_{\dataset, \Vec{x}} \left( | \what_{\gamp}^{\top} \Vec{x} - \what^{\top} \Vec{x} | < \varepsilon \right) \xrightarrow[n, d \to \infty, \sfrac{n}{d} = \alpha]{}_{} 1
\end{equation}

Moreover, we show in \cref{appendix:amp} that in this high-dimensional limit, the estimators $\what_{i, \gamp}$ of~\cref{eq:leave_one_out_from_amp} converge to the true leave-one-out estimators~\cref{eq:argmin_loo}. 

\vspace{-2mm}
\section{Numerical experiments}
\vspace{-2mm}
    
    In this section, we first show that on synthetic Gaussian data, our method correctly approximates the conformity scores while accelerating their computations by orders of magnitude. This allows us to compare FCP to other methods such as split conformal prediction and the Bayes-optimal estimator in a non-trivial high-dimensional setting.
    We then evaluate the methods on real datasets, showing the usefulness of AMP for uncertainty quantification beyond synthetic data with no distributional assumptions.
In all of our numerical experiments, the prediction intervals will have a target coverage of 90\% 

\subsection{Synthetic high-dimensional benchmark}
\vspace{-2mm}
\paragraph{Coverage and size of prediction intervals --} In this section, we consider synthetic data generated by the model described in~\cref{eq:gaussian_data}. In~\cref{tab:length}, we first compute the coverage of \taylorgamp for the Ridge and Lasso regressions at different values of $\lambda$. We see in the right-most column that our method provides the desired coverage. Moreover, on this synthetic data we compare the size of prediction intervals produced by exact LOO and observe that the average length are almost equal. This numerically validates the statement of~\cref{sec:convergence_high_dim} and shows that with Gaussian data, even at moderate dimension, \taylorgamp is very close to exact LOO.

\begin{table*}
    \centering
    \begin{tabular}{c|c|c|c|c||c}
        Problem & exact LOO & \taylorgamp & SCP & CQP & Coverage of \taylorgamp\\
        \toprule
        Lasso ($\lambda = 1$) & 3.9 $\pm$ 0.45 & 4.2 $\pm$ 0.8 & 4.3 $\pm$ 0.9 & 4.7 $\pm$ 0.9 & 0.9 \\
        Ridge ($\lambda = 1$) & 3.7 $\pm$ 0.34 & 3.9 $\pm$ 0.4  & 4.4 $\pm$ 0.8 & 4.7 $\pm$ 0.9  &  0.89 \\
        Ridge ($\lambda = 0.01$) & 4.4 $\pm$ 0.5 & 4.7 $\pm$ 0.7 & 5.7 $\pm$ 1.2 & 4.8 $\pm$ 0.9 & 0.91 \\
    \end{tabular}
    \vspace{3mm}
    \caption{Mean and standard deviation, of the size of prediction intervals at coverage $q = 0.9$, with random data at $n = 100, d = 50$ generated from a Gaussian teacher. For all methods except exact LOO, values are averaged over $1000$ test samples.}
    \label{tab:length}
\end{table*}

\begin{table}
    \centering
    \begin{tabular}{c|c|c}
        Problem & JI (\taylorgamp) & JI (SCP) \\
        \toprule
        Ridge ($\lambda = 0.01$) & 0.93 $\pm$ 0.04 & 0.80 $\pm$ 0.12 \\
        Ridge ($\lambda = 0.1$) & 0.95 $\pm$ 0.04 & 0.83 $\pm$ 0.1 \\
        Ridge ($\lambda = 1$) & 0.98 $\pm$ 0.02 & 0.84 $\pm$ 0.04 \\
        Lasso ($\lambda = 0.01$) & 0.90 $\pm$ 0.06 & 0.86 $\pm$ 0.11 \\
        Lasso ($\lambda = 0.1$) & 0.92 $\pm$ 0.05 & 0.87 $\pm$ 0.09 \\
        Lasso ($\lambda = 1$) & 0.97 $\pm$ 0.03 & 0.88 $\pm$ 0.08 \\
         \bottomrule
     \end{tabular}
     \vspace{3mm}
     \caption{ Jaccard index (JI) between exact LOO and \taylorgamp and SCP for different estimators, with data generated from a Gaussian teacher, and $d = 100, n = 200$. We report the averages and standard deviation  over $20$ test samples.}
    \label{fig:jaccard}
\end{table}

We also compute the similarity between the prediction intervals produced by \taylorgamp with those returned by exact LOO, to show that both methods return the same intervals. To this end, we compute the \textit{Jaccard index} between the exact and approximate intervals. Recall that the Jaccard index between two sets $\interval_1, \interval_2$ is defined as
$$
\jaccard \left( \interval_1, \interval_2 \right) = \frac{| \interval_1 \cap \interval_2|}{|\interval_1 \cup \interval_2|} \in [0, 1]
$$
values closer to $1$ indicate more precise approximations. We report our findings in~\cref{fig:jaccard}, where we evaluate the Jaccard index $\jaccard \left( \interval_{\fcp} (\Vec{x} ), \interval_{\text{\taylorgamp}} (\Vec{x} )\right)$ and $\jaccard \left( \interval_{\fcp} (\Vec{x} ), \interval_{SCP} (\Vec{x} )\right)$. Values are the averaged over $20$ test samples. We observe that \taylorgamp has a higher similarity to FCP than SCP. This confirms that even though our method is an approximation of FCP, even at moderate dimensions it provides intervals that are very close to the exact ones.

\begin{figure}[h]
    \centering
        \centering
         \includegraphics[width=0.49\columnwidth]{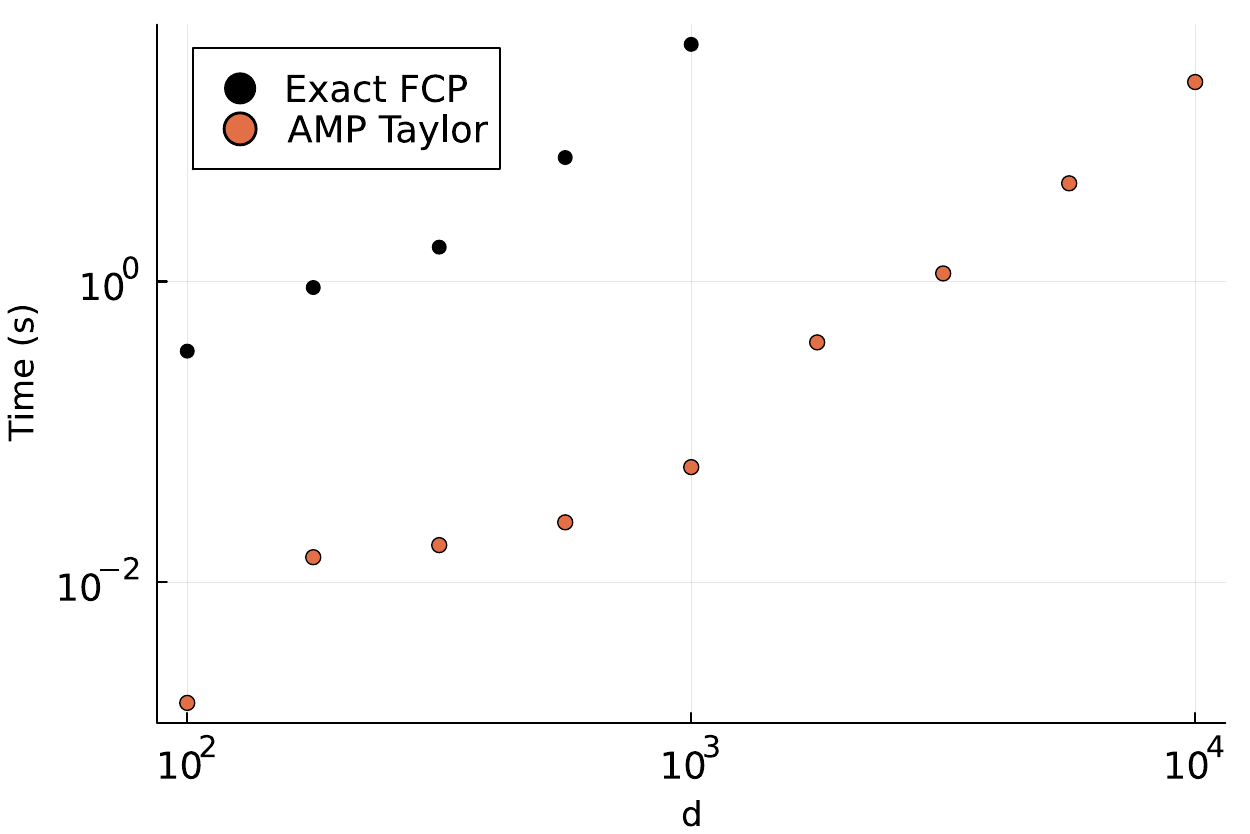}
    \caption{Computation time to produce a single prediction interval, for exact LOO and \taylorgamp, for Lasso at $\lambda = 1$ and $\sfrac{n}{d} = 0.5$.}
     \label{fig:coverage_time_comparison}
 \end{figure}

\paragraph{Computation speed -- } In Figure~\ref{fig:coverage_time_comparison}, we compare the time to compute $\interval ( \Vec{x} )$ for a single test sample $\Vec{x}$, as a function of the dimension for a fixed sampling ratio $\alpha = \sfrac{n}{d}$. We observe that our method provides a speed-up over exact LOO by more than two orders of magnitude, and allows us to quantify the uncertainty for dimensions about 10 times higher for the same amount of time. With the \taylorgamp algorithm, we can readily treat problems of dimension $10^4$. 

To summarize, so far our numerical results show that our algorithm approximates precisely exact LOO, while being order of magnitudes faster. This allows to benchmark FCP against other methods in large dimensions, as we do in the following paragraphs.

\paragraph{Comparison with Bayes posterior --} We compare the prediction intervals of conformal prediction with those of the Bayes-optimal estimator as defined in Section~\ref{sec:def_bayes_optimal}. Recall that the Bayes-optimal estimator has the lowest generalisation error when the data-generating process is known. When the prior $p_{\wstar}$ is Gaussian, the log-posterior exactly corresponds to Ridge regression with $\lambda = 1$. Likewise, for a Laplace prior on $\wstar$, the log-posterior is exactly the empirical risk of Lasso, with $\lambda = 1$. In Table~\ref{tab:fcp_vs_bo}, we compare the average length of the prediction intervals provided by FCP with the highest density intervals of the Bayes posterior. Note that for a Gaussian prior, the posterior distribution is also Gaussian and can be easily sampled. However, this is not the case for a Laplace prior. In general, one would sample the posterior using Monte-Carlo methods. However, within our synthetic data setting, we can leverage the AMP algorithm~\ref{alg:gamp} to sample the posterior~\cite{clarte2023theoretical}. AMP is much faster than costly Monte-Carlo sampling, while being exact in the high-dimensional limit. Lines in bold represent the matched settings where the minimized empirical risk matches the true log posterior. We see that in these settings, FCP has almost optimal length, as it is very close to those of the Bayes-optimal estimator. On the other hand, when $\lambda$ has a value that does not match the true prior, then the intervals obtained with \taylorgamp are significantly larger than those of Bayes, for instance with $\lambda = 0.1$.

\begin{table}[h]
    \centering
    \begin{tabular}{c|c|c|c}
        Teacher & Regularization & Bayes & \taylorgamp \\
        \toprule
        \multirow{3}{*}{Gaussian} & Ridge $(\lambda = 0.1)$ & \multirow{3}{*}{4.4} & 4.9 \\
         & \textbf{Ridge} $(\lambda = 1.0)$ & & \textbf{4.4} \\
         & Lasso $(\lambda = 1.0)$ & & 4.9 \\
        \midrule
        \multirow{3}{*}{Laplace} & Lasso $(\lambda = 0.1)$ & \multirow{3}{*}{5.1} & 13.2 \\
         & \textbf{Lasso} $(\lambda = 1.0)$ & & \textbf{5.6} \\
         & Ridge $(\lambda = 1.0)$ & & 4.9 \\
        \bottomrule
    \end{tabular}
    \vspace{3mm}
    \caption{Average length of prediction intervals of Bayes posterior and FCP with \taylorgamp, at $d = 250, n = 125$. Measures are averaged over $200$ samples of both $\dataset$ and the single test sample. Bold lines correspond to the matched setting where the empirical risk corresponds to the log-posterior of the data-generating process.}
    \label{tab:fcp_vs_bo}
\end{table}

\vspace{-1em}
\paragraph{Comparison with split conformal prediction --} In~\cref{tab:length}, we compare the length of the prediction intervals of \taylorgamp with SCP described in~\ref{eq:simple_scp}, and to conformalized quantile regression (CQP)~\cite{Romano2019Conformalized}, where split conformal prediction is applied on two estimators of the quantile functions of the likelihood $p(y | \vec{x})$. We observe that as expected, our method provides tighter intervals while having the correct coverage.

\vspace{-1em}
\paragraph{Comparison with approximate homotopy--} In this section, we compare the performance of \taylorgamp with \textit{Approximate homotopy} introduced in \cite{ndiaye2019computing}. Like \taylorgamp, the goal of approximate homotopy is to approximate the computation of FCP. In \cref{tab:comparison_with_homotopy_real}, we compare both methods in terms of computation time and coverage, on Lasso regression with $\lambda = 1, \alpha = 0.5$ for a target coverage of $0.9$. For approximate homotopy we used the code provided by the authors of \cite{ndiaye2019computing}
with the default sets of parameters. From the table, we see that at a relatively low dimension, both methods have acceptable runtime but from $d = 150$ onwards, approximate homotopy becomes extremely slow, making it unusable in the high-dimensional regime. 

\begin{table*}[h!]
\centering
\begin{tabular}{c|c|c|c|c|c}
\toprule
\textbf{Dataset} & \textbf{Regularization} & \textbf{Method} & \textbf{Size} & \textbf{Time} & \textbf{Coverage} \\
\toprule
\multirow{2}{*}{Gaussian ($d = 100, n = 1000$)} & \multirow{2}{*}{Lasso ($\lambda = 1$)} & \taylorgamp & 4.4 & 0.007 &  0.89 \\
                            &                              & Approximate homotopy & 4.6 & 0.025 & 0.9 \\
 \midrule
 \multirow{2}{*}{Gaussian ($d = 250, n = 100$)} & \multirow{2}{*}{Lasso ($\lambda = 1$)} & \taylorgamp & 4.6 & 0.044 & 0.9 \\
                            &                              & Approximate homotopy & 4.5 & 10.6 & 0.89 \\
\midrule\midrule
\multirow{3}{*}{Boston} & \multirow{3}{*}{Lasso ($\lambda = 1$)} & AMP & 1.6 & 0.03 & 0.88 \\
 & & \taylorgamp & 9.1 & 0.03 & 0.91 \\
                           &                              & Approximate homotopy & 1.5 & 0.02 & 0.95 \\
\midrule
\multirow{3}{*}{Riboflavin} & \multirow{3}{*}{Lasso ($\lambda = 0.25$)} & AMP & 2.34 & 0.42 & 0.89 \\
 & & \taylorgamp & 9.5 & 0.4 & 0.95 \\
                           &                              & Approximate homotopy & 0.23 & 17.7 & 1.0 \\
\bottomrule
\end{tabular}
\caption{Comparison of \taylorgamp and approximate homotopy on synthetic (Top) and real (Bottom) datasets. We observe that at high dimension, \taylorgamp is faster by several orders of magnitude, while providing the correct coverage. }
\label{tab:comparison_with_homotopy_real}
\end{table*}
\subsection{Real data}
We also compare the performance of both AMP and \taylorgamp with approximate homotopy on real data. We use for this two datasets, the Boston housing dataset and Riboflavin production data\cite{Buhlmann2014HighDimensional}. We validate that our methods provide the correct coverage. 

In particular, we observe that our method significantly speeds up FCP compared to approximate homotopy on the riboflavin dataset, where $d = 4088$, in exchange of wide prediction intervals. On the other hand, we observe that \taylorgamp behaves poorly as the prediction intervals are much wider than the two other methods.

Note that the convergence of AMP heavily depends on the properties of the input matrix $X$, making it somewhat fragile for real data. However, in cases in which it converges such as the ones shown here, the coverage is competitive. 

\vspace{-1em}
\section{Discussion}

In this paper, we introduce a method to accelerate the computations of full conformal prediction while guaranteeing confidence sets with the correct coverage. Our method leverages methods stemming from high-dimensional statistics literature, namely the approximate message passing (AMP) algorithm. 
Our numerical experiments on synthetic and real data show that the method has the potential to provide narrow confidence sets (with coverage guaranteed) while reducing the computation time by almost three orders of magnitude compared to the baseline. 
Our method has a particular theoretical interest, as \taylorgamp can be used to investigate more easily the properties of full conformal prediction in high dimensions by drastically speeding up the simulations. 
The proposed algorithm can leverage the fact that it is asymptotically exact on the synthetic Gaussian data and these data can thus be used as a benchmark for other speed-up methods in high-dimensions. 


\paragraph{Possible extensions --} In~\cref{appendix:classification}, we describe how to extend our method to classification, and leave a more detailed study to future work. While we only investigated conformal prediction for frequentist estimators, AMP can be used to sample from Bayesian posteriors more efficiently than Monte-Carlo methods. Our results could thus be extended to Bayesian conformal prediction, where the conformity scores are given by the predictive posterior~\cite{Fong2021ConformalBayesian}. Moreover, the computation of leave-one-out estimators could be applied to other methods such as the Jackknife+\cite{Barber2019Predictive}. Finally, we did not leverage the state-evolution equations of AMP, which allow us to characterize precisely the behaviour of AMP in high dimensions. This theoretical study is left to future work. 

\paragraph{Limitations --} One limitation of our work is the assumption weak dependence on every sample in \taylorgamp. Further, while we show that our method is applicable to real data, the stability of AMP heavily depends on the input matrix. The extension of our method to more complex algorithms of a similar kind such as VAMP~\cite{Rangan2019Vector}, which would make our method applicable to a broader set of data, and is left to future work.

The code used to produce the figures can be found in the following github repository: \url{https://github.com/lclarte/ConformalAmp.jl}. All experiments were run on a Apple M1 Pro laptop with 16 Go of memory. 

\section{Acknowledgements}
We thank Bruno Loureiro and Florent Krzakala for valuable discussions. This research was supported by the NCCR MARVEL, a National Centre of Competence in Research, funded by the Swiss National Science Foundation (grant number 205602). 

\clearpage

\bibliography{bibliography}

\begin{thebibliography}{39}
\providecommand{\natexlab}[1]{#1}
\providecommand{\url}[1]{\texttt{#1}}
\expandafter\ifx\csname urlstyle\endcsname\relax
  \providecommand{\doi}[1]{doi: #1}\else
  \providecommand{\doi}{doi: \begingroup \urlstyle{rm}\Url}\fi

\bibitem[Angelopoulos and Bates(2022)]{angelopoulos2022gentle}
Anastasios~N. Angelopoulos and Stephen Bates.
\newblock A gentle introduction to conformal prediction and distribution-free uncertainty quantification, 2022.

\bibitem[Angelopoulos et~al.(2021)Angelopoulos, Bates, Cand{\`e}s, Jordan, and Lei]{angelopoulos2021learn}
Anastasios~N Angelopoulos, Stephen Bates, Emmanuel~J Cand{\`e}s, Michael~I Jordan, and Lihua Lei.
\newblock Learn then test: Calibrating predictive algorithms to achieve risk control.
\newblock \emph{arXiv preprint arXiv:2110.01052}, 2021.

\bibitem[Arpino et~al.(2024{\natexlab{a}})Arpino, Liu, and Venkataramanan]{arpino2024inferring}
Gabriel Arpino, Xiaoqi Liu, and Ramji Venkataramanan.
\newblock Inferring change points in high-dimensional linear regression via approximate message passing.
\newblock In Ruslan Salakhutdinov, Zico Kolter, Katherine Heller, Adrian Weller, Nuria Oliver, Jonathan Scarlett, and Felix Berkenkamp, editors, \emph{Proceedings of the 41st International Conference on Machine Learning}, volume 235 of \emph{Proceedings of Machine Learning Research}, pages 1841--1864. PMLR, 21--27 Jul 2024{\natexlab{a}}.

\bibitem[Arpino et~al.(2024{\natexlab{b}})Arpino, Liu, and Venkataramanan]{arpino24Inferring}
Gabriel Arpino, Xiaoqi Liu, and Ramji Venkataramanan.
\newblock Inferring change points in high-dimensional linear regression via approximate message passing.
\newblock In Ruslan Salakhutdinov, Zico Kolter, Katherine Heller, Adrian Weller, Nuria Oliver, Jonathan Scarlett, and Felix Berkenkamp, editors, \emph{Proceedings of the 41st International Conference on Machine Learning}, volume 235 of \emph{Proceedings of Machine Learning Research}, pages 1841--1864. PMLR, 21--27 Jul 2024{\natexlab{b}}.

\bibitem[Bai et~al.(2021{\natexlab{a}})Bai, Mei, Wang, and Xiong]{Bai2021Dont}
Yu~Bai, Song Mei, Haiquan Wang, and Caiming Xiong.
\newblock Don't just blame over-parametrization for over-confidence: Theoretical analysis of calibration in binary classification.
\newblock In \emph{International Conference on Machine Learning}, 2021{\natexlab{a}}.

\bibitem[Bai et~al.(2021{\natexlab{b}})Bai, Mei, Wang, and Xiong]{Bai2021Understanding}
Yu~Bai, Song Mei, Huan Wang, and Caiming Xiong.
\newblock Understanding the under-coverage bias in uncertainty estimation.
\newblock In \emph{Neural Information Processing Systems}, 2021{\natexlab{b}}.

\bibitem[Barber et~al.(2019)Barber, Cand{\`e}s, Ramdas, and Tibshirani]{Barber2019Predictive}
Rina~Foygel Barber, Emmanuel~J. Cand{\`e}s, Aaditya Ramdas, and Ryan~J. Tibshirani.
\newblock Predictive inference with the jackknife+.
\newblock \emph{The Annals of Statistics}, 2019.

\bibitem[Bayati and Montanari(2010)]{Bayati2010Dynamics}
Mohsen Bayati and Andrea Montanari.
\newblock The dynamics of message passing on dense graphs, with applications to compressed sensing.
\newblock \emph{2010 IEEE International Symposium on Information Theory}, pages 1528--1532, 2010.

\bibitem[B{\"u}hlmann et~al.(2014)B{\"u}hlmann, Kalisch, and Meier]{Buhlmann2014HighDimensional}
Peter B{\"u}hlmann, Markus Kalisch, and Lukas Meier.
\newblock High-dimensional statistics with a view toward applications in biology.
\newblock 2014.

\bibitem[Cherubin et~al.(2021)Cherubin, Chatzikokolakis, and Jaggi]{cherubin2021exact}
Giovanni Cherubin, Konstantinos Chatzikokolakis, and Martin Jaggi.
\newblock Exact optimization of conformal predictors via incremental and decremental learning.
\newblock In \emph{International Conference on Machine Learning}, 2021.

\bibitem[Clarté et~al.(2023{\natexlab{a}})Clarté, Loureiro, Krzakala, and Zdeborova]{clarte2022overparametrized}
Lucas Clarté, Bruno Loureiro, Florent Krzakala, and Lenka Zdeborova.
\newblock On double-descent in uncertainty quantification in overparametrized models.
\newblock In Francisco Ruiz, Jennifer Dy, and Jan-Willem van~de Meent, editors, \emph{Proceedings of The 26th International Conference on Artificial Intelligence and Statistics}, volume 206 of \emph{Proceedings of Machine Learning Research}, pages 7089--7125. PMLR, 2023{\natexlab{a}}.

\bibitem[Clarté et~al.(2023{\natexlab{b}})Clarté, Loureiro, Krzakala, and Zdeborová]{clarte2023theoretical}
Lucas Clarté, Bruno Loureiro, Florent Krzakala, and Lenka Zdeborová.
\newblock Theoretical characterization of uncertainty in high-dimensional linear classification.
\newblock \emph{Machine Learning: Science and Technology}, 4\penalty0 (2):\penalty0 025029, jun 2023{\natexlab{b}}.

\bibitem[Clarté et~al.(2024)Clarté, Vandenbroucque, Dalle, Loureiro, Krzakala, and Zdeborová]{clarte2024analysis}
Lucas Clarté, Adrien Vandenbroucque, Guillaume Dalle, Bruno Loureiro, Florent Krzakala, and Lenka Zdeborová.
\newblock Analysis of bootstrap and subsampling in high-dimensional regularized regression.
\newblock In \emph{Uncertainty in Artificial Intelligence}, 2024.

\bibitem[Davison and Hinkley(1997)]{Davison1997Bootstrap}
A.~C. Davison and D.~V. Hinkley.
\newblock \emph{Bootstrap Methods and their Application}.
\newblock Cambridge Series in Statistical and Probabilistic Mathematics. Cambridge University Press, 1997.

\bibitem[Depope et~al.(2024)Depope, Mondelli, and Robinson]{Depope2024Inference}
Al~Depope, Marco Mondelli, and Matthew~R. Robinson.
\newblock Inference of genetic effects via approximate message passing.
\newblock In \emph{ICASSP 2024 - 2024 IEEE International Conference on Acoustics, Speech and Signal Processing (ICASSP)}, pages 13151--13155, 2024.

\bibitem[Donoho et~al.(2009)Donoho, Maleki, and Montanari]{Donoho2009Message}
David~L. Donoho, Arian Maleki, and Andrea Montanari.
\newblock Message-passing algorithms for compressed sensing.
\newblock \emph{Proceedings of the National Academy of Sciences}, 106\penalty0 (45):\penalty0 18914–18919, November 2009.
\newblock ISSN 1091-6490.

\bibitem[El~Alaoui et~al.(2022)El~Alaoui, Montanari, and Sellke]{el2022sampling}
Ahmed El~Alaoui, Andrea Montanari, and Mark Sellke.
\newblock Sampling from the sherrington-kirkpatrick gibbs measure via algorithmic stochastic localization.
\newblock In \emph{2022 IEEE 63rd Annual Symposium on Foundations of Computer Science (FOCS)}, pages 323--334. IEEE, 2022.

\bibitem[Fong and Holmes(2021)]{Fong2021ConformalBayesian}
Edwin Fong and Chris~C Holmes.
\newblock Conformal bayesian computation.
\newblock In M.~Ranzato, A.~Beygelzimer, Y.~Dauphin, P.S. Liang, and J.~Wortman Vaughan, editors, \emph{Advances in Neural Information Processing Systems}, volume~34, pages 18268--18279. Curran Associates, Inc., 2021.

\bibitem[Jing et~al.(2018)Jing, Max, Alessandro, Ryan, and Larry]{Lei2018Distribution}
Lei Jing, G'Sell Max, Rinaldo Alessandro, J.~Tibshirani Ryan, and Wasserman Larry.
\newblock Distribution-free predictive inference for regression.
\newblock \emph{Journal of the American Statistical Association}, 113\penalty0 (523):\penalty0 1094--1111, 2018.

\bibitem[Karoui and Purdom(2018)]{Karoui2018Can}
Noureddine~El Karoui and Elizabeth Purdom.
\newblock Can we trust the bootstrap in high-dimensions? the case of linear models.
\newblock \emph{J. Mach. Learn. Res.}, 19:\penalty0 5:1--5:66, 2018.

\bibitem[Lei(2017)]{lei2017fast}
Jing Lei.
\newblock Fast exact conformalization of lasso using piecewise linear homotopy.
\newblock In \emph{Biometrika}, 2017.

\bibitem[Liang and Sur(2022)]{Liang2022Precise}
Tengyuan Liang and Pragya Sur.
\newblock A precise high-dimensional asymptotic theory for boosting and minimum-l1-norm interpolated classifiers.
\newblock \emph{The Annals of Statistics}, 50\penalty0 (3), 2022.

\bibitem[Loureiro et~al.(2023)Loureiro, Gerbelot, Refinetti, Sicuro, and Krzakala]{Loureiro2023Fluctuations}
Bruno Loureiro, C{\'e}dric Gerbelot, Maria Refinetti, Gabriele Sicuro, and Florent Krzakala.
\newblock Fluctuations, bias, variance and ensemble of learners: exact asymptotics for convex losses in high-dimension.
\newblock \emph{Journal of Statistical Mechanics: Theory and Experiment}, 2023, 2023.

\bibitem[Millard et~al.(2020)Millard, Hess, Mailh{\'{e}}, and Tanner]{Millard2020MRI}
Charles Millard, Aaron~T. Hess, Boris Mailh{\'{e}}, and Jared Tanner.
\newblock An approximate message passing algorithm for rapid parameter-free compressed sensing {MRI}.
\newblock In \emph{{IEEE} International Conference on Image Processing, {ICIP} 2020, Abu Dhabi, United Arab Emirates, October 25-28, 2020}, pages 91--95. {IEEE}, 2020.

\bibitem[Mézard and Montanari(2009)]{Mezard2009Information}
Marc Mézard and Andrea Montanari.
\newblock \emph{{Information, Physics, and Computation}}.
\newblock Oxford University Press, 01 2009.
\newblock ISBN 9780198570837.

\bibitem[Ndiaye and Takeuchi(2019)]{ndiaye2019computing}
Eugene Ndiaye and Ichiro Takeuchi.
\newblock Computing full conformal prediction set with approximate homotopy.
\newblock volume~32. Curran Associates, Inc., 2019.

\bibitem[Nixon and Tran(2020)]{Nixon2020WhyA}
Jeremy Nixon and Dustin Tran.
\newblock Why aren’t bootstrapped neural networks better?
\newblock Neurips 2020 ICBINB Workshop, 2020.

\bibitem[Obuchi and Kabashima(2016)]{Obuchi2016Cross}
Tomoyuki Obuchi and Yoshiyuki Kabashima.
\newblock Cross validation in lasso and its acceleration.
\newblock \emph{Journal of Statistical Mechanics: Theory and Experiment}, 2016\penalty0 (5):\penalty0 053304, may 2016.

\bibitem[Papadopoulos et~al.(2002)Papadopoulos, Proedrou, Vovk, and Gammerman]{Papadopoulos2002Inductive}
Harris Papadopoulos, Kostas Proedrou, Volodya Vovk, and Alex Gammerman.
\newblock Inductive confidence machines for regression.
\newblock pages 345--356, 2002.

\bibitem[Pearl(1988)]{Pearl2014Probabilistic}
Judea Pearl.
\newblock \emph{Probabilistic Reasoning in Intelligent Systems: Networks of Plausible Inference}.
\newblock Morgan Kaufmann Publishers Inc., San Francisco, CA, USA, 1988.
\newblock ISBN 1558604790.

\bibitem[Quenouille(1956)]{Quenouille1956Notes}
M.~H. Quenouille.
\newblock Notes on bias in estimation.
\newblock \emph{Biometrika}, 43\penalty0 (3/4):\penalty0 353--360, 1956.

\bibitem[Rangan et~al.(2019)Rangan, Schniter, and Fletcher]{Rangan2019Vector}
Sundeep Rangan, Philip Schniter, and Alyson~K. Fletcher.
\newblock Vector approximate message passing.
\newblock \emph{IEEE Transactions on Information Theory}, 65\penalty0 (10):\penalty0 6664--6684, 2019.

\bibitem[Romano et~al.(2019)Romano, Patterson, and Candes]{Romano2019Conformalized}
Yaniv Romano, Evan Patterson, and Emmanuel Candes.
\newblock Conformalized quantile regression.
\newblock In H.~Wallach, H.~Larochelle, A.~Beygelzimer, F.~d\textquotesingle Alch\'{e}-Buc, E.~Fox, and R.~Garnett, editors, \emph{Advances in Neural Information Processing Systems}, volume~32. Curran Associates, Inc., 2019.

\bibitem[Shafer and Vovk(2007)]{shafer2007tutorial}
Glenn Shafer and Vladimir Vovk.
\newblock A tutorial on conformal prediction, 2007.

\bibitem[Sur and Candès(2019)]{Sur2019Modern}
Pragya Sur and Emmanuel~J. Candès.
\newblock A modern maximum-likelihood theory for high-dimensional logistic regression.
\newblock \emph{Proceedings of the National Academy of Sciences}, 116\penalty0 (29):\penalty0 14516–14525, July 2019.
\newblock ISSN 1091-6490.

\bibitem[Takahashi(2024)]{takahashi2024replica}
Takashi Takahashi.
\newblock A replica analysis of under-bagging, 2024.

\bibitem[Vovk et~al.(2005)Vovk, Gammerman, and Shafer]{vovk1005algorithmic}
Vladimir Vovk, Alex Gammerman, and Glenn Shafer.
\newblock \emph{Algorithmic Learning in a Random World}.
\newblock 01 2005.

\bibitem[Zdeborová and Krzakala(2016)]{Zdeborova2016Statistical}
Lenka Zdeborová and Florent Krzakala.
\newblock Statistical physics of inference: thresholds and algorithms.
\newblock \emph{Advances in Physics}, 65\penalty0 (5):\penalty0 453–552, August 2016.
\newblock ISSN 1460-6976.

\bibitem[Zhu et~al.(2006)Zhu, Rosset, Zou, and Hastie]{Zhu2006Multiclass}
Ji~Zhu, Saharon Rosset, Hui Zou, and Trevor Hastie.
\newblock Multi-class adaboost.
\newblock \emph{Statistics and its interface}, 2, 02 2006.

\end{thebibliography}


\clearpage
\appendix

\onecolumn

\section{Approximate Message Passing to approximate leave-one-out residuals}
\label{appendix:amp}

\subsection{Introduction of relaxed-Belief Propagation and Approximate Message Passing}


In this section, we explain how AMP can be used to compute the leave-one-out residuals used in~\cref{eq:scores_fcp}. The naive way to compute these residuals is to fit the leave-one-out estimators $\what_{-i} ( y ) $ for each sample $1 \leqslant i \leqslant n$ and each possible label $y$, which requires $n \times | \mathcal{Y} |$ fits, with $\mathcal{Y}$ the set of candidate labels, typically a discretization of $\mathbb{R}$. We will first see that AMP can be used to compute all the $\what_{-i}$

To introduce AMP, we first consider the following problem. Consider a dataset $\mathcal{D} = \left( \Vec{x}_i, y_i \right)_{i = 1}^n$ of size $n$.
Assume that the data is generated from the model~\eqref{eq:gaussian_data}, where the input $\vec{x}_i \in \mathbb{R}^d$ are sampled according to $\mathcal{N}(\Vec{0}, \sfrac{I_d}{d})$, and the labels are generated from a \textit{teacher} as $y \sim p(y | \wstar^{\top} \vec{x})$. Our goal is to sample the following distribution 
\begin{equation}
    p(\vectheta) = \frac{1}{Z} \prod_{i = 1}^n P_{out} \left( y_i | \vectheta^{\top} \Vec{x}_i \right) \prod_{\mu = 1}^d P_{\theta}( \vectheta_{\mu} ) 
    \label{eq:def_distribution_amp}
\end{equation}
The empirical risk minimization problem~\eqref{eq:def_erm} introduced in~\cref{sec:setting} is a particular instance of~\cref{eq:def_distribution_amp} where 
\begin{equation}
    P_{out}(y | z) \propto e^{- \beta \ell(y, z)}, \qquad P_{\theta}( z ) \propto e^{- \beta r(z)}
    \label{eq:probas_for_erm}
\end{equation}
in the limit $\beta \to \infty$. The starting point of approximate message passing is the writing of the belief-propagation algorithm for the graph associated with~\cref{eq:def_distribution_amp}, where the variable-nodes of the graph are the coordinates $\vectheta_{\mu}$ and the factor nodes, representing the interaction between the variable-nodes, are the observations $y_i$. The message passing consists in iterating messages $m_{\mu \to i}$ from variable to factor-nodes and $m_{i \to \mu}$ from factor to variable-nodes. These messages read
\begin{align}
    m_{\mu \to i} ( \theta_{\mu} ) &= \frac{1}{z_{i \to \mu}} P_{\theta} ( \theta_{\mu}) \prod_{j \neq i} m_{j \to \mu} ( \theta_{\mu} )  \\ 
    m_{i \to \mu} ( \theta_{\mu} ) &= \frac{1}{z_{\mu \to i}} \int \prod_{\nu \neq \mu} \dd \theta_{\nu} m_{\nu \to i} P_{out} \left( y_i | \sum_{\nu} \Vec{x}_{i \nu} \theta_{\nu} \right)
    \label{eq:def_messages_bp}
\end{align}
This messages give access to the distribution $p\left( \vec{\theta} \right)$ and in particular this marginals : indeed, the marginal distribution $p(\theta_{\mu})$ is given by 
\begin{equation}
    p(\theta_{\mu}) = \frac{1}{z_{\mu}} P_{\theta}(\theta_{\mu}) \prod_{i = 1}^n m_{i \to \mu} (\theta_{\mu}) 
\end{equation}
where $z_{\mu}$ is a normalization constant. Iterating~\cref{eq:def_messages_bp} is not tractable, especially in high-dimensions as it involves $(d - 1)$ integrals to update each $m_{i \to \mu}$. To make these equations tractable, one can use relaxed-Belief Propagation (rBP), which relies on the central limit theorem and the projection of the messages on their first two moments. We thus define the \textit{cavity mean} $\hat{\theta}_{\mu \to i}$ and \textit{cavity variance} $\hat{v}_{\mu \to i}$ as 
\begin{align}
    \hat{\theta}_{\mu \to i} &= \int \dd \theta_{\mu} \theta_{\mu} m_{\mu \to i} ( \theta_{\mu} ) \\
    \hat{v}_{\mu \to i}      &= \int \dd \theta_{\mu} \theta_{\mu}^2 m_{\mu \to i} ( \theta_{\mu} ) - \hat{\theta}^2_{\mu \to i}
\end{align}
In particular, the vector $\left( \hat{\theta}_{\mu \to i} \right)_{\mu = 1}^d$ represents the mean of the marginals of distribution~\eqref{eq:def_distribution_amp} in the absence of the $i$-th sample. In the context of empirical risk minimization, this is exactly the leave-one-out estimator $\what_{-i}$ defined as 
\begin{equation}
    \what_{-i} = \arg\min_{\Vec{\theta}} \sum_{j \neq i} \ell \left( y_j, \Vec{\theta}^{\top} \Vec{x}_j \right) + \sum_{\mu = 1}^d r(\theta_{\mu})
\end{equation}
Our goal is thus to compute efficiently the cavity means and use them to compute the leave-one-out residuals. 

\paragraph{rBP} The main idea behind rBP is to iteratively compute the cavity means and variances, to obtain the desired marginal mean and variance of $\Vec{\theta}$. We define $\omega_{i \to \mu}, V_{i \to \mu}$ the mean and variance of the messages $m_{i \to \mu}$ and $\hat{\theta}_{\mu \to i}, \hat{v}_{\mu \to i}$ the mean and variance of $m_{\mu \to i}$. 

We detail rBP in~\cref{alg:rBP}, and refer to~\cite[Chapter VI, Section C]{Zdeborova2016Statistical} for a detailed explanation of the algorithm. In particular, the algorithm makes use of the \textit{channel} and \textit{denoising} functions $\channel$ and $\denoiser$ functions, defined respectively as 
\begin{equation}
    \channel(y, \omega, V) = \frac{\partial \log \mathcal{Z}_y (y, \omega, V)}{\partial \omega}, \qquad \mathcal{Z}_y (y, \omega, V) = \int \dd z P_{out} (y | z) e^{- \frac{1}{2V}(z - \omega)^2}
    \label{eq:def_channel}
\end{equation}
and 
\begin{equation}
    \denoiser(b, A) = \frac{\partial \log \mathcal{Z}_w (b, A)}{\partial b}, \qquad \mathcal{Z}_w (b, A) = \int \dd x P_{\theta} (x) e^{bx - \frac{A}{2}x^2} 
    \label{eq:def_denoiser}
\end{equation}

In the case of empirical risk minization~\eqref{eq:def_erm}, using the prior and likelihood from \cref{eq:probas_for_erm} into the definitions\eqref{eq:def_channel} and \eqref{eq:def_denoiser} and taking the limit $\beta \to \infty$ yields Equation~\eqref{eq:def_channel_denoiser}.

\paragraph{From rBP to AMP} Note that in rBP, we iterate over $n \times d$ means and variances $\omega_{i \to \mu}, V_{i \to \mu}, \hat{\theta}_{\mu \to i}, \hat{v}_{\mu \to i}$, which scales quadratically with the dimension in the high-dimensional limit where $n, d \to \infty$ with a constant sampling ratio $\sfrac{n}{d} = \alpha$. However, a key observation is that the quantities $\hat{\theta}_{\mu \to i}, \hat{v}_{\mu \to i}$ only weakly depend on $\mu$, and similarly $\omega_{i \to \mu}, V_{i \to \mu}$ weakly depend on $\mu$. Hence, let us define 
\begin{align}
    \begin{cases}
        \omega_{i} &= \sum_{\mu} \Vec{x}_{i\mu} \hat{\theta}_{\mu \to i}  \\
        V_{i} &= \sum_{\mu} \Vec{x}_{i\mu}^2 \hat{v}_{\mu \to i}
    \end{cases},
    \qquad
    \begin{cases}
        A_{\mu} &= - \sum_{i = 1}^n \partial_{\omega} g_{out} \left( y_i, \omega_i, V_i \right) \vec{x}^2_{i \mu} \\
        b_{\mu} &= \sum_{i = 1}^n g_{out} \left( y_i, \omega_{i \to \mu}, V_{i \to \mu} \right) \vec{x}_{i \mu} \\
    \end{cases}
\end{align}
note that for all $\mu$ and all $i$, in the high-dimensonal limit considered here we have 
\begin{align}
    \omega_i &= \omega_{i \to \mu} + \Vec{x}_{i\mu} \hat{\theta}_{\mu \to i} = \omega_{i \to \mu} + O \left( \sfrac{1}{\sqrt{n}}\right) \\
    V_i      &= V_{i \to \mu} + \vec{x}_{i \mu}^2 \hat{v}_{\mu \to i} =  V_{i \to \mu} + O\left( \sfrac{1}{n} \right)
\end{align}

As a consequence, we have for all $\mu$ and all $i$
\begin{align}
     A_{\mu} &= - \sum_{j = 1}^n \vec{x}^2_{j \mu} \partial_{\omega} g_{out} \left( y_j, \omega_{j}, V_{j} \right) = \sum_{j = 1}^n \vec{x}^2_{j \mu} \left[ \partial_{\omega} g_{out} \left( y_j, \omega_{j \to \mu}, V_{j \to \mu} \right) + O(\sfrac{1}{\sqrt{n}}) \right] \\
     &= - \sum_{j = 1}^n \vec{x}^2_{j \mu} \partial_{\omega} g_{out} \left( y_j, \omega_{j \to \mu}, V_{j \to \mu} \right) + O(\sfrac{1}{\sqrt{n}}) \\
    &= - \sum_{j \neq i}^n \vec{x}^2_{j \mu} \partial_{\omega} g_{out} \left( y_j, \omega_{j \to \mu}, V_{j \to \mu} \right) + O(\sfrac{1}{\sqrt{n}}) \\
     &= - A_{\mu \to i} +  O \left( \sfrac{1}{\sqrt{n}} \right)
 \end{align}
Similarly, we get 
 \begin{align}
     b_{\mu} &= b_{\mu \to i} + O \left( \sfrac{1}{\sqrt{n}} \right) 
 \end{align}

So that one can simply compute the estimator $\vec{\theta} = \denoiser \left( \vec{b}, \vec{A} \right)$. The challenge is to compute the vectors $\vec{\omega}, \vec{V}, \vec{b}$. To do so, we note that 
\begin{align}
    \channel \left( y_i, \omega_{i \to \mu}, V_{i \to \mu} \right) &= \channel \left( y_i, \omega_{i}, V_{i} \right) - \vec{x}_{i \mu} \hat{\theta}_{\mu \to i} \partial_{\omega} \channel \left( y_i, \omega_{i \to \mu}, V_{i \to \mu} \right) + O \left( \sfrac{1}{n} \right) \\
\end{align}
such that 
\begin{align}
    b_{\mu} &= \sum_{i = 1}^n \vec{x}_{i \mu} \channel \left( y_i, \omega_{i}, V_{i} \right) - \sum_i \vec{x}^2_{i \mu} \hat{\theta}_{\mu} \partial_{\omega} \channel \left( y_{i}, \omega_i, V_i \right) + O \left( \sfrac{1}{\sqrt{n}} \right) \\
\end{align}

Moreover, 
\begin{align}
    \omega_i &= \sum_{\mu = 1}^d \Vec{x}_{i\mu} \hat{\theta}_{\mu \to i} = \sum_{\mu} \Vec{x}_{i\mu} \left( \hat{\theta}_{\mu} - \vec{x}_{i \mu} v_{\mu} \channel \left( y_i, \omega_i, V_i \right) \right) + O(\sfrac{1}{n} )\\
\end{align}

These iterative equations are, in the leading order, the same as those shown in~\cref{alg:gamp}. In the high-dimensional regime, these iteratives coincide with rBP. Going from rBP to AMP, we have reduced the number of variables to iterate on from $O(n \times d)$ to $O(n + d)$, and can still recover the marginal distribution by 
\begin{equation}
    \hat{\theta}_{\mu} = \denoiser \left( b_{\mu}, A_{\mu} \right)
\end{equation}

\subsection{Recovering the leave-one-out estimators from AMP}

For each sample $i$, computing the leave-one-out estimator $\what_{-i}$ means computing the marginals of the distribution
\begin{equation}
    p(\vec{\theta}) = \frac{1}{Z} \prod_{j \neq i} P_{out} \left( y_j | \vec{\theta}^{\top} \vec{x}_j \right) \prod_{\mu = 1}^d P_{\theta} \left( \vec{\theta}_{\mu} \right)
    \label{eq:def_distribution_loo}
\end{equation}
with $P_{out}$ and $P_{\theta}$ defined in~\cref{eq:probas_for_erm} and where the sample $(\vec{x}_i, y_i)$ is removed from the data. Our method leverages the fact that these marginals are computed iteratively by relaxed-BP and stored in the variables $\hat{\theta}_{\mu \to i}$. Indeed, each $\hat{\theta}_{\mu \to i}$ stores the posterior mean of $\theta_{mu}$ when the interaction node $i$ is removed from the graph associated to~\cref{eq:def_distribution_amp}, which corresponds exactly to the distribution of~\cref{eq:def_distribution_loo}. While rBP explicitly computes these quantities, its computational complexity makes it unusable. Instead, we will recover these estimators from AMP. Indeed, at the leading order we have : 

\begin{align}
    \hat{\theta}_{\mu \to i} &= \denoiser \left( b_{\mu \to i}, A_{\mu \to i} \right) = \denoiser \left( b_{\mu \to i}, A_{\mu} \right) + O \left( \sfrac{1}{n} \right) \\
                &=  \denoiser \left( b_{\mu}, A_{\mu} \right) - b_{i \to \mu} \partial_b \denoiser \left( b_\mu, A_\mu \right) + O \left( \sfrac{1}{n} \right) = \what_\mu - \channel (y_i, \omega_i, V_i) \Vec{x}_{i \mu} \vhat_\mu + O \left( \sfrac{1}{n} \right)
    \label{eq:from_marginal_to_cavity_mean}
\end{align}

The expression on the right-hand side corresponds to the approximation of the leave-one-out estimators $\what_{-i, \gamp}$ used in~\cref{alg:gamp}. 

\paragraph{Convergence of the leave-one-out residuals in high-dimensions} Under the assumptions~\eqref{eq:gaussian_data}, we see from~\cref{eq:from_marginal_to_cavity_mean} thatin the high-dimensional limit the leave-one-out estimators computed by AMP will converge to the exact ones at a $O(\sfrac{1}{n})$ rate. As such, for a given test sample $\vec{x}, y$ the approximated residuals $y - \vec{x}^{\top} \vec{\theta}_{-i, \gamp}$ will converge to $y - \vec{x}^{\top} \vec{\theta}_{-i}$ at a $O \left( \sfrac{1}{\sqrt{n}} \right)$ rate. This implies that asymptotically the prediction intervals built using the AMP leave-one-out converge to the prediction intervals with the exact residuals.

\paragraph{Applying AMP without Gaussian assumptions} 
We thus see that from AMP, we get an approximation of the leave-one-out estimator that can be used to compute the residuals in~\cref{eq:scores_fcp}. The derivations performed in this section were done under the assumption that the input data are Gaussian with i.i.d. covariance and $\sfrac{1}{d}$ variance. However, AMP can be applied on any data, with no guarantee a priori on its performance.

\begin{algorithm}[tb]
    \caption{relaxed-Belief Propagation}
    \label{alg:rBP}
    \begin{algorithmic}

    \REPEAT
        \STATE {\bfseries Input:} Dataset $\dataset = \left( \Vec{x}_i, y_i \right)_{i = 1}^n$

        \STATE \begin{align}
            \begin{cases}
                V_{i \to \mu}^{t}      &= \sum_{\nu \neq \mu} \vec{x}^2_{i \mu} v_{\nu \to i}^{t-1} \\
                \omega_{i \to \mu}^{t} &= \sum_{\nu \neq \mu} \vec{x}_{i \mu} \hat{\theta}_{\nu \to i}^{t-1}
            \end{cases}   
        \end{align}

        \STATE \begin{align}
        \begin{cases}
            A_{\mu \to i}^{t} &= - \sum_{j \neq i} \partial_{\omega} \channel \left( y_j, \omega_{j \to \mu}^t, V_{j \to \mu} \right) \Vec{x}_{j \mu}^2 \\
            b_{\mu \to i}^{t} &= \sum_{j \neq i} \channel \left( y_j, \omega_{j \to \mu}^t, V_{j \to \mu} \right) \Vec{x}_{j \mu}
        \end{cases}
        \end{align}

        \STATE \begin{align}
         \hat{\theta}_{\mu \to i}^{t} &= \denoiser \left( b_{\mu \to i}^{t}, A_{\mu \to i}^{t} \right) \\
         \hat{v}_{\mu \to i}^{t}      &= \partial_b \denoiser \left( b_{\mu \to i}^{t}, A_{\mu \to i}^{t} \right)
        \end{align}
        
    \UNTIL{Convergence  of $\hat{\theta}_{\mu \to i}, \hat{v}_{\mu \to i}$}
        \STATE {\bfseries Return} $\what, \vhat$ \textbf{ such that :}  
        \begin{align}
            \hat{\theta}_{\mu} &= \denoiser \left( \sum_i b_{\mu \to i}, \sum_i A_{\mu \to i} \right) \\
            \hat{v}_{\mu} &= \partial_b \denoiser \left( \sum_i b_{\mu \to i}, \sum_i A_{\mu \to i} \right)
        \end{align} 
    \end{algorithmic}
\end{algorithm}

\section{Derivation ot \taylorgamp}
\label{appendix:taylorgamp}

In this section, we derive the \taylorgamp algorithm. Our starting point is AMP, derived in Appendix~\ref{appendix:amp}. In what follow, we consider a dataset $\mathcal{D}$ of size $n+1$ to stay consistent with the notation of the main text. Our goal is to compute the variation of the $\what_{-i}$ to the first order with respect to the last label $y_{n+1}$. To this end, we will write the vectors defined in AMP $\what(y), \vhat(y), \Vec{g}(y), \partial\Vec{g}(y), \Vec{b}(y), \vec{A}(y), \Vec{\omega}(y), \Vec{V}(y)$ as functions of $y_{n+1} = y$  

For the sake of conciseness, let us define the vector 
\begin{equation}
    \Omega \left( y \right) = \left( \what(y), \vhat(y), \Vec{\omega}(y), \Vec{V}(y), \Vec{g}(y), \partial\Vec{g}(y), \Vec{b}(y), \vec{A}(y) \right) \in \mathbb{R}^{4 \times (d + n)}
    \label{eq:def_omega}
\end{equation}

Then, $\Omega (y)$ is the fixed point of the equation 
$$
\Omega (y) = \fgamp ( \Omega(y), y)
$$
where the function $\fgamp(\Omega) = \left( f_{\gamp}^{\what}, f_{\gamp}^{\vhat}, f_{\gamp}^{\Vec{\omega}}, f_{\gamp}^{\Vec{V}}, f_{\gamp}^{\Vec{g}}, f_{\gamp}^{\partial \Vec{g}}, f_{\gamp}^{\Vec{b}}, f_{\gamp}^{\partial \Vec{A}} \right)$ is defined as 
\begin{align}
    \begin{cases}
        \fgamp^{\what} &= \denoiser(\Vec{b}, \vec{A})\\
        \fgamp^{\vhat} &= \partial_b \denoiser(\Vec{b}, \vec{A})\\
        \fgamp^{\Vec{\omega}} &= X\what - \Vec{V} \odot \Vec{g} \\
        \fgamp^{\Vec{V}} &= X^2 \vhat \\
        \fgamp^{\Vec{g}} &= \channel \left( \Vec{y}, \Vec{\omega}, \Vec{V} \right) \\
        \fgamp^{\partial \Vec{g}} &= \partial_{\omega} \channel \left( \Vec{y}, \Vec{\omega}, \Vec{V} \right) \\
        \fgamp^{\Vec{b}} &= X^{\top} \Vec{g} + \Vec{A} \odot \what \\
        \fgamp^{\partial \Vec{A}} &= - X^{2 \top} \partial \Vec{g} \\
    \end{cases}
    \label{eq:def_f_gamp}
\end{align}

Equivalently, we have $\Omega(y) - \fgamp(\Omega (y), y) = \Vec{0}$. Under the assumption that the function $\Omega(y)$ is differentiable, one can use the implicit function theorem around a value $\hat{y}$ to write 
\begin{align}
    \frac{\partial \Omega}{\partial y}\left( \hat{y} \right)                 &= \left( \mathbf{I} - Jac \left( \fgamp \right) \right)^{-1} \frac{\partial \fgamp}{\partial y} \left( \hat{y} \right) \\
    \Leftrightarrow \frac{\partial \Omega}{\partial y}\left( \hat{y} \right) &=  Jac \left( \fgamp \right) \left( \frac{\partial \Omega}{\partial y}\left( \hat{y} \right) \right) + \frac{\partial \fgamp}{\partial y} \left( \hat{y} \right)
\end{align}

From the last equality we find that we can compute the derivative $\frac{\partial \Omega}{\partial y}\left( \hat{y} \right)$ by iterating the following system of linear equations over a vector $\Delta \Omega^t$ :
\begin{equation}
    \Delta \Omega^{t+1} =  Jac \left( \fgamp \right) \left( \Delta \Omega^{t} \right) + \frac{\partial \fgamp}{\partial y} \left( \hat{y} \right)
    \label{eq:iteration_delta_omega}
\end{equation}

The jacobian of the function $\fgamp$ is written 
\begin{align}
    \begin{cases}
        Jac f_{\amp}^{\what}            &= \left( \Vec{0}, \Vec{0}, \Vec{0}, \Vec{0}, \partial_b \denoiser(\Vec{b}, \Vec{A}), \partial_A \denoiser(\Vec{b}, \Vec{A}) \right) \\
        Jac f_{\amp}^{\vhat}            &= \left( \Vec{0}, \Vec{0}, \Vec{0}, \Vec{0}, \partial_b \partial_b \denoiser(\Vec{b}, \Vec{A}), \partial_A \partial_b \denoiser(\Vec{b}, \Vec{A}) \right) \\
        Jac f_{\amp}^{\omega}           &= \left( X, \Vec{0}, \Vec{0}, - Diag(\Vec{g}), - Diag(\Vec{V}), \Vec{0}, \Vec{0}, \Vec{0} \right) \\
        Jac f_{\amp}^{\Vec{V}}          &= \left( \vec{0}, X^2, \Vec{0}, \Vec{0}, \Vec{0}, \Vec{0} \right) \\
        Jac f_{\amp}^{\Vec{g}}          &= \left( \vec{0}, \vec{0}, \partial_{\omega} \Vec{g}, \partial_{V} \Vec{g}, \Vec{0}, \Vec{0}, \Vec{0}, \Vec{0} \right) \\
        Jac f_{\amp}^{\partial \Vec{g}} &= \left( \vec{0}, \vec{0}, \partial_{\omega} \partial_{\omega} \Vec{g}, \partial_{V} \partial_{\omega} \Vec{g}, \Vec{0}, \Vec{0} \right) \\
        Jac f_{\amp}^{\Vec{b}}          &= \left( Diag(\Vec{A}), \Vec{0}, \vec{0}, \vec{0}, X^{\top}, \Vec{0}, \Vec{0}, Diag(\Vec{w}) \right) \\
        Jac f_{\amp}^{\Vec{A}}          &= \left( \vec{0}, \Vec{0}, \Vec{0}, \Vec{0}, - X^{2\top}, \Vec{0},\Vec{0}, \Vec{0} \right)\\
    \end{cases}
    \label{eq:jacobian_f_gamp}
\end{align}

and the derivative $\frac{\partial \fgamp}{\partial y}$ with respect to the last label is 
\begin{align*}
    \begin{cases}
        \partial_y f_{\amp}^{\Vec{\theta}} &= \Vec{0} \\
        \partial_y f_{\amp}^{\Vec{v}} &= \Vec{0} \\
        \partial_y f_{\amp}^{\Vec{\omega}} &= \Vec{0}\\
        \partial_y f_{\amp}^{\Vec{V}} &= \Vec{0} \\
        \partial_y f_{\amp}^{\Vec{g}} &= \left( 0, \cdots, 0, \partial_y g(y_n, \omega_n, V_n) \right) \\
        \partial_y f_{\amp}^{\partial \Vec{g}} &= \left( 0, \cdots, 0, \partial_y \partial_{\omega} g(y_n, \omega_n, V_n) \right) \\
        \partial_y f_{\amp}^{\Vec{b}} &= \Vec{0} \\
        \partial_y f_{\amp}^{\Vec{A}} &= \Vec{0} \\
    \end{cases}
\end{align*}

When writing Equation~\eqref{eq:iteration_delta_omega} with the expression of the Jacobian of Equation~\eqref{eq:jacobian_f_gamp}, one obtains the iterations of \taylorgamp in Algorithm~\ref{alg:gamp_order_one}.

\subsection{Justification of \taylorgamp}
\label{appendix:taylorgamp_justification}

\begin{figure}
    \centering
    \includegraphics[width=0.45\textwidth]{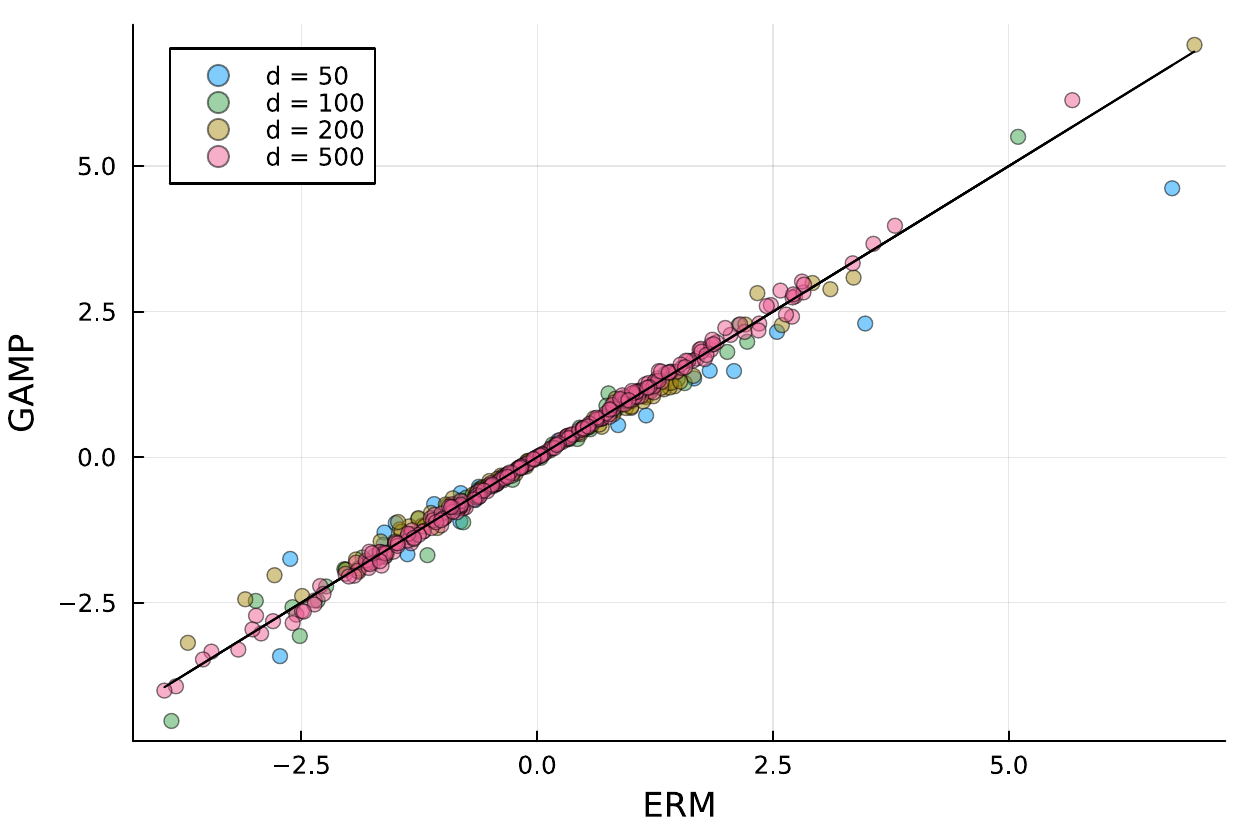}
    \includegraphics[width=0.45\textwidth]{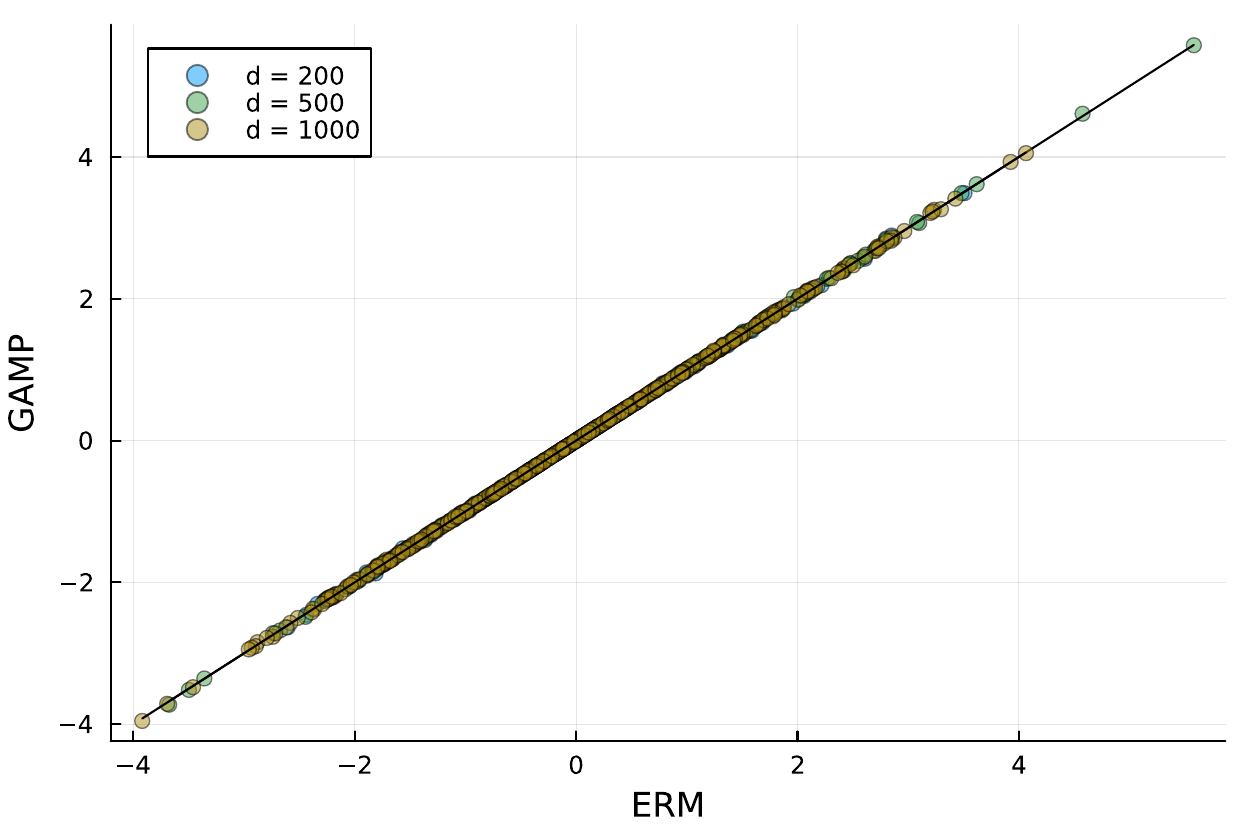}
    \includegraphics[width=0.45\textwidth]{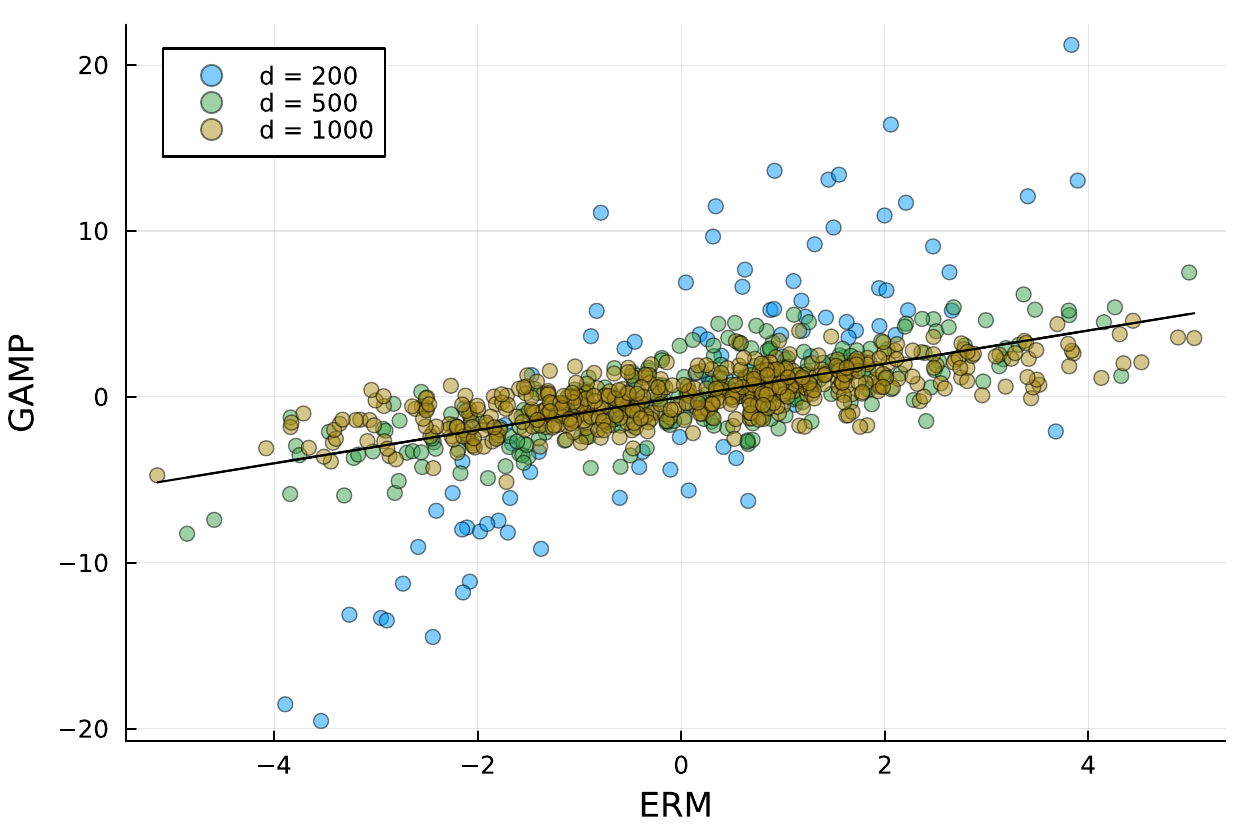}
    \includegraphics[width=0.45\textwidth]{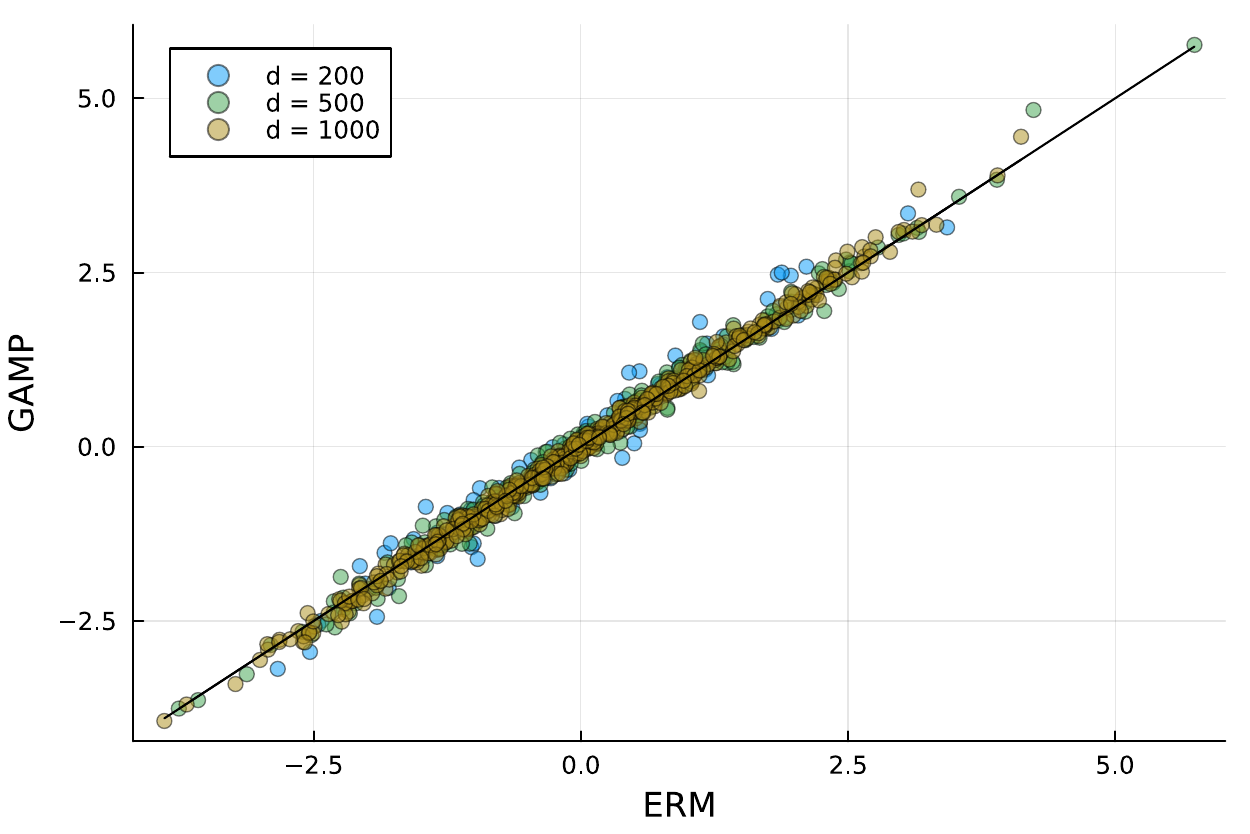}
    \caption{Comparison of the leave-one-out estimators computed exactly by solving~\cref{eq:argmin_loo} and by \taylorgamp, for Ridge (top row) and Lasso (bottom row), as $\lambda = 0.01$ (left column) and $\lambda = 1$ (right column). All plots are at $\sfrac{n}{d} = 0.5$}
    \label{fig:residuals_gamp_erm}
\end{figure}

As stated in the previous subsection, \taylorgamp is based on the assumption that the function $y \to \Omega \left) y \right)$ is differentiable. Our underlying assumption behind \taylorgamp is that the leave-one-out residuals only weakly depend on the last label in high-dimensions. We numerically justify this assumption in~\cref{fig:residuals_gamp_erm}. In this Figure, we compare the leave-one-out residuals obtained by computing the estimators $\what_{-i}$ exactly and with \taylorgamp for different settings. To do so, we sample a dataset $\dataset$ at random. We use~\cref{alg:gamp} and \cref{alg:gamp_order_one} to compute the $\what_{-i, \gamp}(y_n)$ and $\Delta \what_{-i, \gamp}(y)$ as prescribed above. Then, we change the last label $y_n \to y_n + \delta y$ with $\delta y = 5$. After this change we compute the leave-one-estimators exactly $\what_{-i} (y_n + \delta y)$ and use our linear approximation $\what_{-i, \gamp} (y + \delta y) = \what_{-i} ( y ) + \delta y \Delta \what_{-i. \gamp} (y)$. We then compare $\what_{-i} (y_n + \delta y)^{\top} \vec{x}_i$ and our approximation $\what_{-i, \gamp} (y_n + \delta y)^{\top} \vec{x}_i$ that is used to compute our conformity scores. As we observe in the figure, at high dimensions $d = 1000$, our approximations are very close to the true values, meaning that \taylorgamp will accurately estimate the scores (hence the prediction intervals) of FCP.

We note however from the lower-left plot that at moderate dimension, \taylorgamp does not precisely approximates the leave-one-out residuals for the LASSO, which partly explains the mediocre results obtained by \taylorgamp on real data in~\cref{tab:comparison_with_homotopy_real} in the main.

\section{Coverage guarantee for AMP}
\label{appendix:amp_coverage}

First, we show that AMP is symmetric : indeed, consider a permutation $s : [1, n] \rightarrow [1, n]$ and $S$ the corresponding permutation matrix defined as $S_{ij} = \delta \left( j = s(i) \right)$. Then, consider running AMP on the permutated data $\Tilde{X} = SX$ and labels $\Tilde{y} = SY$. At each iteration $t$, the channel vectors $\Tilde{\Vec{g}}^t, \Tilde{\partial{\Vec{g}}}^t$ 
$\Tilde{\Vec{g}}^t = S \Vec{g}^t$ and $\Tilde{\partial \Vec{g}^t} = S \partial \Vec{g}^t$. Then, the vectors $\Vec{b}^t, \Vec{A}^t$ now become 
\begin{align}
\begin{cases}
\Tilde{\vec{A}}^t &= - \Tilde{X^{2\top}} \Tilde{\partial \vec{g}^t} = - X^{2\top} S^T S \Vec{g}^t = \Vec{A}^t \\
\Tilde{\vec{b}}^t &= \Tilde{X^{\top}} \Tilde{\vec{g}^t} + \Tilde{\Vec{A}^t} \otimes \what^t = X^{\top} S^T S \vec{g}^t + \Tilde{\Vec{A}^t} \otimes \what^t = \Vec{b}^t \\
\end{cases}
\end{align}
and by recursion we deduce that the estimator of AMP $(\what, \vhat)$ given after convergence is invariant under permutation. Then, the scores computed from~\cref{eq:leave_one_out_from_amp} are symmetric. Then, under the assumption that the data $(\vec{x}_i, y_i)$ is exchangeable, we obtain~\cref{prop:coverage} : in expectation over the training and test data
\begin{equation}
    \mathbb{P}_{\mathcal{D}, \vec{x}} \left( y \in \interval(\vec{x})  \right) \geqslant 1 - \kappa
\end{equation}

\section{Details on real datasets}

In this section, we provide details on the datasets used in~\cref{tab:comparison_with_homotopy_real}. We use the Boston housing dataset (availa containing 506 samples at dimension 14, and the Riboflavin dataset~\cite{Buhlmann2014HighDimensional} of 71 samples at dimension 4088. Both datasets were normalized and split randomly into training and test data, with 80 percent of the data being in the training set.
The coverage, average time and interval size are obtained by computing the average on the test set, and averaging over $20$ random train / test splits. 

For~\cref{tab:comparison_with_homotopy_real}, approximate homotopy was used with the default parameters provided by the authors. The same data preprocessing was applied as for our method to compare fairly the size of the prediction intervals.

\section{Full conformal prediction for classification}
\label{appendix:classification}

In the main part of the paper, we focused on the case of regression, however conformal prediction has been successfully applied for classification tasks~\cite{angelopoulos2021learn, angelopoulos2022gentle}. Consider a classification task with $k$ classes, where a predictor estimate the probabilities $p_1 (\vec{x} ), \cdots, p_n ( \vec{x} )$. Then, the conformity scores are defined as 
\begin{equation}
    \score_i = \sum_{k = 1}^{\pi^{-1}(y)} p_{\pi(1)}
    \label{eq:score_classification}
\end{equation}
where $\pi$ is a permutation that ranks the classes by decreasing order of probability, i.e $p_{\pi(1)} > \cdots > p_{\pi(K)}$. In words, the score is the sum of the probability of all the classes whose $p_i$ is higher of equal to the true observed class.

\paragraph{Binary classification} In the context of generalized linear model, one might train an estimator using the cross entropy loss with an $L_2$ regularizer. For $K = 2$ classes, this is logistic regression 
\begin{equation}
    \what = \arg\min_{\vectheta} - \sum_{i = 1}^n \log \left( 1 + e^{- y_i \times \vec{x}_i^{\top} \vectheta} \right) + \sfrac{\lambda}{2} \| \vectheta \|^2
\end{equation}

As for regression, one can use AMP and \taylorgamp with the adequate channel and denoising function to estimate $\what$, and compute the leave-one-out estimators using~\cref{eq:leave_one_out_from_amp}. For the logistic loss, the channel function is defined as 
\begin{align}
    \channel(y, \omega, V) = \frac{{\rm prox} \ell_{\omega, V}(y, \cdot) - \omega}{V}, \qquad {\rm prox} \ell_{\omega, V}(y, \cdot) = \arg\min_z \ell(y, z) + \frac{1}{2V} \left( z - \omega \right)^2
\end{align}

\clearpage

\end{document}